\documentclass[10pt,journal,compsoc]{IEEEtran}
 \pdfoutput=1
\ifCLASSOPTIONcompsoc
  \usepackage[nocompress]{cite}
\else
  \usepackage{cite}
\fi

\usepackage{acronym}
\usepackage{hyperref}
\usepackage{multirow}
\usepackage{multicol}
\usepackage{graphicx}
\usepackage{amsmath,amssymb}
\usepackage{verbatim}

\usepackage{ragged2e}
\usepackage{subcaption}
\usepackage{grffile}
\usepackage{cleveref}
\usepackage{pgfplots}
\usepackage{color,soul}

\definecolor{blue_1a}{RGB}{93,133,195}
\definecolor{green_3a}{RGB}{80,182,149}
\definecolor{red_8a}{RGB}{239,85,85}
\definecolor{org_2a}{RGB}{250,136,70}
\definecolor{yellow_a}{RGB}{204,204,70}

\definecolor{blu}{RGB}{11,29,120}
\definecolor{pur}{RGB}{0,105,192}
\definecolor{pin}{RGB}{0,169,181}
\definecolor{lre}{RGB}{0,198,152}
\definecolor{lblu}{RGB}{87,160,211}

\acrodef{AE}[AE]{\emph{Autoencoder}}
\acrodef{FPR}[FPR]{\emph{False Positive Rate}}

\acrodef{AUC}[AUC]{\emph{Area Under the Curve}}

\acrodef{ROC}[ROC]{\emph{Receiver Operating Characteristic}}

\acrodef{TPR}[TPR]{\emph{True Positive Rate}}
\acrodef{MSE}[MSE]{\emph{Mean Square Error}}
\acrodef{SOTA}[SOTA]{\emph{State of the Art}}
\acrodef{GT}[GT]{\emph{Geometric Transformations}}
\acrodef{VAE}[VAE]{\emph{Variational Autoencoder}}
\acrodef{GAN}[GAN]{\emph{Generative Adversarial Network}}
\acrodef{ARAE}[ARAE]{\emph{Adversarial Robust Trained Autoencoder}}
\ifCLASSINFOpdf
\else
\fi
\begin{document}
%
% paper title
% Titles are generally capitalized except for words such as a, an, and, as,
% at, but, by, for, in, nor, of, on, or, the, to and up, which are usually
% not capitalized unless they are the first or last word of the title.
% Linebreaks \\ can be used within to get better formatting as desired.
% Do not put math or special symbols in the title.
\title{Puzzle-AE: Novelty Detection in Images through Solving Puzzles}
%
%
% author names and IEEE memberships
% note positions of commas and nonbreaking spaces ( ~ ) LaTeX will not break
% a structure at a ~ so this keeps an author's name from being broken across
% two lines.
% use \thanks{} to gain access to the first footnote area
% a separate \thanks must be used for each paragraph as LaTeX2e's \thanks
% was not built to handle multiple paragraphs
%
%
%\IEEEcompsocitemizethanks is a special \thanks that produces the bulleted
% lists the Computer Society journals use for "first footnote" author
% affiliations. Use \IEEEcompsocthanksitem which works much like \item
% for each affiliation group. When not in compsoc mode,
% \IEEEcompsocitemizethanks becomes like \thanks and
% \IEEEcompsocthanksitem becomes a line break with idention. This
% facilitates dual compilation, although admittedly the differences in the
% desired content of \author between the different types of papers makes a
% one-size-fits-all approach a daunting prospect. For instance, compsoc 
% journal papers have the author affiliations above the "Manuscript
% received ..."  text while in non-compsoc journals this is reversed. Sigh.

\author{\IEEEauthorblockN{Mohammadreza Salehi, Ainaz Eftekhar\textsuperscript{*}, Niousha Sadjadi\textsuperscript{*}, Mohammad Hossein Rohban, Hamid R. Rabiee}
\IEEEauthorblockA{Department of Computer Engineering\\Sharif University of Technology \\
Tehran, Iran. \\ Email: smrsalehi@ce.sharif.edu,
aeftekhar@ce.sharif.edu, nsadjadi@ce.sharif.edu,
rohban@sharif.edu, rabiee@sharif.edu\\
\textsuperscript{*}Denotes equal contribution}}

% note the % following the last \IEEEmembership and also \thanks - 
% these prevent an unwanted space from occurring between the last author name
% and the end of the author line. i.e., if you had this:
% 
% \author{....lastname \thanks{...} \thanks{...} }
%                     ^------------^------------^----Do not want these spaces!
%
% a space would be appended to the last name and could cause every name on that
% line to be shifted left slightly. This is one of those "LaTeX things". For
% instance, "\textbf{A} \textbf{B}" will typeset as "A B" not "AB". To get
% "AB" then you have to do: "\textbf{A}\textbf{B}"
% \thanks is no different in this regard, so shield the last } of each \thanks
% that ends a line with a % and do not let a space in before the next \thanks.
% Spaces after \IEEEmembership other than the last one are OK (and needed) as
% you are supposed to have spaces between the names. For what it is worth,
% this is a minor point as most people would not even notice if the said evil
% space somehow managed to creep in.

% The paper headers
\markboth{Journal of \LaTeX\ Class Files,~Vol.~14, No.~8, August~2015}%
{Shell \MakeLowercase{\textit{et al.}}: Bare Demo of IEEEtran.cls for Computer Society Journals}
% The only time the second header will appear is for the odd numbered pages
% after the title page when using the twoside option.
% 
% *** Note that you probably will NOT want to include the author's ***
% *** name in the headers of peer review papers.                   ***
% You can use \ifCLASSOPTIONpeerreview for conditional compilation here if
% you desire.

% The publisher's ID mark at the bottom of the page is less important with
% Computer Society journal papers as those publications place the marks
% outside of the main text columns and, therefore, unlike regular IEEE
% journals, the available text space is not reduced by their presence.
% If you want to put a publisher's ID mark on the page you can do it like
% this:
%\IEEEpubid{0000--0000/00\$00.00~\copyright~2015 IEEE}
% or like this to get the Computer Society new two part style.
%\IEEEpubid{\makebox[\columnwidth]{\hfill 0000--0000/00/\$00.00~\copyright~2015 IEEE}%
%\hspace{\columnsep}\makebox[\columnwidth]{Published by the IEEE Computer Society\hfill}}
% Remember, if you use this you must call \IEEEpubidadjcol in the second
% column for its text to clear the IEEEpubid mark (Computer Society jorunal
% papers don't need this extra clearance.)

% use for special paper notices
%\IEEEspecialpapernotice{(Invited Paper)}

% for Computer Society papers, we must declare the abstract and index terms
% PRIOR to the title within the \IEEEtitleabstractindextext IEEEtran
% command as these need to go into the title area created by \maketitle.
% As a general rule, do not put math, special symbols or citations
% in the abstract or keywords.
\IEEEtitleabstractindextext{%
\begin{abstract}
As an essential part of many anomaly detection methods, the autoencoder lacks flexibility on normal data in complex datasets. U-Net is proven effective for this purpose but overfits the training data if trained only using reconstruction error similar to other AE-based frameworks. As a pretext task of self-supervised learning (SSL) methods, Puzzle-solving has earlier proved its ability in learning semantically meaningful features. We show that training U-Nets based on this task effectively prevents overfitting and facilitates learning beyond pixel-level features. Shortcut solutions, however, are a big challenge in SSL tasks, including jigsaw puzzles. We propose robust adversarial training as an effective automatic shortcut removal. We achieve competitive or superior results compared to the \ac{SOTA} anomaly detection methods on various toy and real-world datasets. Unlike many competitors, the proposed framework is stable, fast, data-efficient, and does not require unprincipled early stopping.
\end{abstract}

% Note that keywords are not normally used for peerreview papers.
}

% make the title area
\maketitle

% To allow for easy dual compilation without having to reenter the
% abstract/keywords data, the \IEEEtitleabstractindextext text will
% not be used in maketitle, but will appear (i.e., to be "transported")
% here as \IEEEdisplaynontitleabstractindextext when the compsoc 
% or transmag modes are not selected <OR> if conference mode is selected 
% - because all conference papers position the abstract like regular
% papers do.
\IEEEdisplaynontitleabstractindextext
% \IEEEdisplaynontitleabstractindextext has no effect when using
% compsoc or transmag under a non-conference mode.

% For peer review papers, you can put extra information on the cover
% page as needed:
% \ifCLASSOPTIONpeerreview
% \begin{center} \bfseries EDICS Category: 3-BBND \end{center}
% \fi
%
% For peerreview papers, this IEEEtran command inserts a page break and
% creates the second title. It will be ignored for other modes.
\IEEEpeerreviewmaketitle

% --------------------Introduction-----------------
\IEEEraisesectionheading{\section{Introduction}\label{sec:introduction}}
% Computer Society journal (but not conference!) papers do something unusual
% with the very first section heading (almost always called "Introduction").
% They place it ABOVE the main text! IEEEtran.cls does not automatically do
% this for you, but you can achieve this effect with the provided
% \IEEEraisesectionheading{} command. Note the need to keep any \label that
% is to refer to the section immediately after \section in the above as
% \IEEEraisesectionheading puts \section within a raised box.

% The very first letter is a 2 line initial drop letter followed
% by the rest of the first word in caps (small caps for compsoc).
% 
% form to use if the first word consists of a single letter:
% \IEEEPARstart{A}{demo} file is ....
% 
% form to use if you need the single drop letter followed by
% normal text (unknown if ever used by the IEEE):
% \IEEEPARstart{A}{}demo file is ....
% 
% Some journals put the first two words in caps:
% \IEEEPARstart{T}{his demo} file is ....
% 
% Here we have the typical use of a "T" for an initial drop letter
% and "HIS" in caps to complete the first word.
\IEEEPARstart{A}{nomaly/novelty} is defined as any digression from the essential features of any given phenomenon. The main task of novelty detection is to infer deviated features from extracted normal training samples’ features. For instance, having a model trained on healthy brain ct-scan images, it should be able to find non-healthy test input images by comparing current extracted features, and the expected ones with different metrics \cite{chalapathy2019deep,chen2018unsupervised,baur2018deep}.

\begin{figure}[!ht]
     \includegraphics[width=\linewidth]{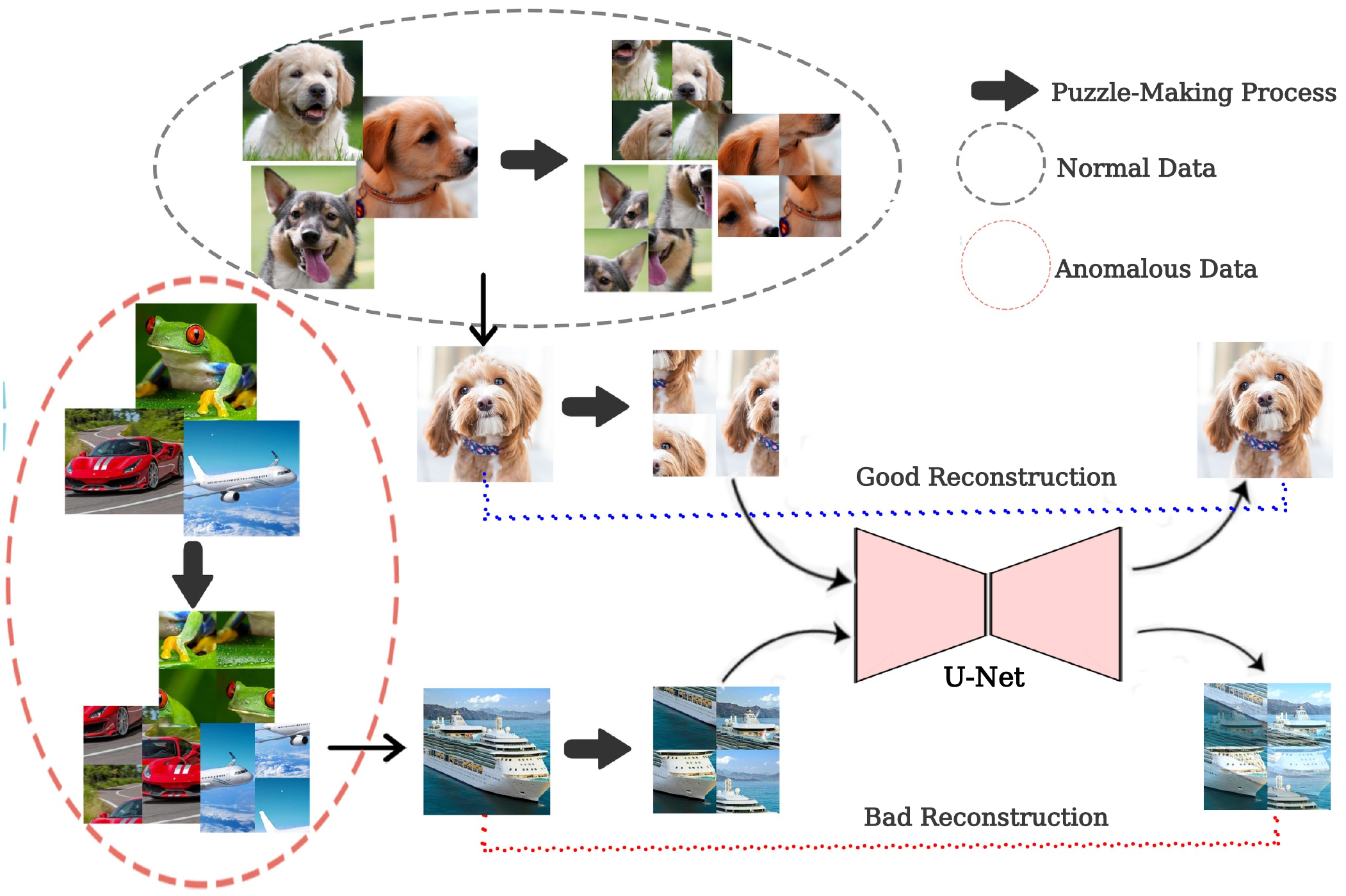}
      \caption{
    %   Reconstruction of LSA \cite{abati2019latent}, MemAE \cite{gong2019memorizing} and Puzzle-AE on normal and anomalous test inputs. Models are trained on the given class. Last row shows the reconstruction of an anomalous input when the models are trained on class Automobile as the normal class. Obviously our model produces significantly better normal outputs while destroys the anomalous ones.
    Reconstruction of normal and anomalous inputs during the testing phase. As it is shown, the model is unable to solve the puzzle for anomalous inputs, which do not have the main features existing in the normal data. As a result, anomalous samples produce high reconstruction loss, whereas normal inputs have low reconstruction loss since their puzzles are perfectly solved by the model.}
      \label{Abstract}
\end{figure}

Although \ac{AUC} has been used as the primary distinctive metric between binary classifiers’ performances, this criterion is not sufficient alone. That is because \ac{AUC} shows the average performance of a model in different operating points. However, a fixed operating point of the \ac{ROC} curve is needed in many practical applications, which is usually when \ac{TPR} is equal to 0.99 or 0.995. In some applications, such as medical and industrial defect detections, data efficiency and real-time performance are key factors for practicality \cite{chalapathy2019deep,gong2019memorizing}. Another practical criterion is the adaptability of a given framework to other datasets. For example, we generally require the model to be agnostic to various hyperparameters' choices and have access to well-known criteria to determine when to stop training of the model. In this paper, we propose a new framework and compare its practicality with the \ac{SOTA} approaches concerning the mentioned criteria. Our results show this framework performs well under all these criteria.

In the literature, GAN based methods \cite{goodfellow2016nips},\cite{kodali2017convergence} suffer from non-reproducibility of the results \cite{salimans2016improved},\cite{martin2017towards} and data hungriness. Likewise, autoregressive approaches have significantly poor performance \cite{nalisnick2018deep}. Moreover, we show that one-class methods such as DSVDD \cite{ruff2018deep} suffer from converging to trivial solutions and need confronting techniques such as early stopping that is not easy to find due to the one-class formulation of the problem. These all cause the generality and impracticality problem.

On the other hand, AE-based methods often have a straightforward training process and yield reproducible results. It has been observed that an \ac{AE} that is trained on just normal samples can reconstruct the normal test inputs while failing to reconstruct anomalous test samples \cite{sakurada2014anomaly}. However, various kinds of \ac{AE}s  have the issue of low-quality input reconstruction on complex datasets such as CIFAR-10 \cite{cifar}. The reason for this phenomenon should be investigated in the training procedure of \ac{AE}, which tries to model pixel intensities with a complex function that is represented by the decoder and leads to
% Modeling the input image at the pixel level could deprive the \ac{AE} of a more general abstraction, which means
finding fallacious relationships between unnecessary or irrelevant features. 

More recently, self-supervised learning methods have shown great potential to go beyond pixel-level and learn semantically meaningful features. Impressive unsupervised classification accuracy on the ImageNet \cite{deng2009imagenet} dataset in \cite{wu2018unsupervised},\cite{oord2018representation},\cite{zhang2017split} has attracted a lot of attention to this field. GT \cite{golan2018deep} has introduced the first one-class classification method that utilizes self-supervised learning with great performance on the CIFAR-10 dataset \cite{cifar}. 
% Despite promising results of the mentioned work on CIFAR-10, we show that in addition to the noted deficiencies of one-class classification methods, in many instances, their performance is not even as good as the base \ac{AE} on real-world datasets such as MVTecAD \cite{bergmann2019mvtec} and Medical images\cite{kitamura2018hemorrhage}\cite{chakrabarty2019tumor}. 
However, we show that their performance is not even as good as the base \ac{AE} on real-world datasets such as MVTecAD \cite{bergmann2019mvtec} and Medical images \cite{kitamura2018hemorrhage,chakrabarty2019tumor}. 

We show that one way to benefit from all the significant aspects of \ac{AE}s and alleviate their deficiencies is using U-Net \cite{ronneberger2015u} as a highly expressive \ac{AE}-based model mixed with a well-defined and unambitious SSL pretext task. That is because U-Net has shown its ability in high-resolution image tasks such as segmentation \cite{badrinarayanan2017segnet,litjens2017survey}. However, it could easily overfit when used similar to previous \ac{AE} based methods such as denoising, etc. One excellent SSL pretext task that solves the overfitting problem and has a minor effect on the data agnosticism of \ac{AE}s, since it keeps almost all the input information, is a puzzle-solving task \cite{noroozi2016unsupervised,hendrycks2019using}. Training a U-Net to solve 4-part puzzled inputs like Fig. \ref{Abstract}, would preserve good abilities of \ac{AE}s while learning how to model normal input data in patch-level rather than pixel one. 

However, according to \cite{noroozi2016unsupervised}, puzzles could be solved easily by finding low-level statistic shortcuts such as edge positions or patches’ mean and variance. Instead of using traditional approaches such as manual jittering, adversarial robust training \cite{madry2017towards,ilyas2019adversarial} as an automatic shortcut removal is used to make the shortcuts out of access for the U-Net, which enhances semantical abstraction modeling. Finally, having trained the framework in a GAN-based setting by considering the U-Net as the generator and adding a discriminator, we could increase the quality of generated images even more \cite{sabokrou2018adversarially}.

Our framework shows great performance on a large number of datasets. To the best of our knowledge, our work is the first study that produces a framework {\it without any need to design unprincipled early stopping criterion} while producing stable and reproducible results. Our main contribution is to provide solutions for the following issues: 
\begin{enumerate}[noitemsep]
\item {\bf High-quality normal sample reconstruction}: Significantly improving the \ac{AE} flexibility to better reconstruct the normal samples in complex real-world datasets. This is achieved by learning beyond pixel-level abstraction using self-supervised training, removing some of the shortcut features by robust adversarial training, and improving the quality of generated images by training the whole framework similar to the \ac{GAN}.

\item {\bf Method Stability}:  The generative nature of our proposed self-supervised method helps in reaching a stable model across the training epochs. We empirically notice the lack of this property in the earlier self-supervised frameworks for anomaly detection. 

\item {\bf Shortcuts in the Jigsaw}: Relieving the shortcut problem of the jigsaw puzzle pretext task automatically by robust adversarial training (which is traditionally solved by manual jittering).

\item {\bf Method Evaluation}: Introducing new practical criteria, \ac{FPR} at a high \ac{TPR}, and robustness to the test-time adversarial attacks that have not been tested on any recent \ac{SOTA} models. 

\item {\bf Method Generality}: Competitive or better than \ac{SOTA} performance on a {\it wide} range of problems without any need of unprincipled early stopping, and with a stable training process, yielding robust and reproducible results.
\end{enumerate}
\begin{figure*}[t!]
\centering
    \includegraphics[width=\textwidth]{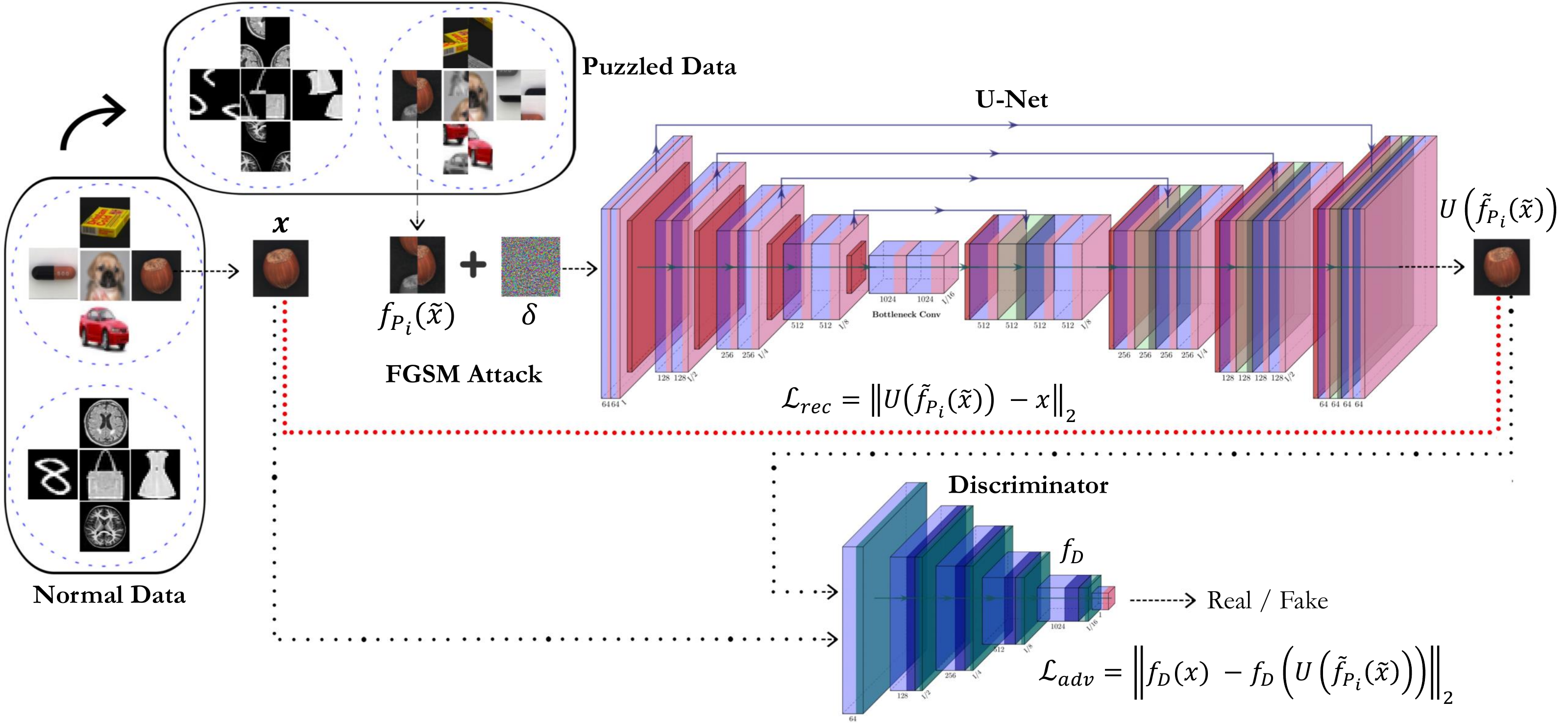}
      \caption{Anomaly Detection framework. As it illustrates, puzzled inputs are added to an FGSM noise to prevent shortcut detection. Then they are given to the U-Net to be solved. The whole framework is trained similar to GAN to improve the quality of reconstructed images.}
       \label{method}
\end{figure*}

% -------------------Related Works-----------------
\section{Related Works}

The major approaches to novelty detection are \ac{AE}-based and one-class classification methods. Latent space autoregression (LSA) \cite{abati2019latent} is a popular \ac{AE}-based method, which fits an autoregressive model to the \ac{AE} bottleneck layer. By jointly training an autoregressive and \ac{AE} model, it can learn a compact latent space for the normal samples. Hence, anomalous samples would have high reconstruction errors, but the value of their bottleneck layer would have a lower probability than the normal ones. This probability is called the ``surprise score" in LSA. At testing time, the surprise score is added to the mean reconstruction error, and the sum is then thresholded to determine the class of a given sample. 

OCGAN \cite{perera2019ocgan} uses an \ac{AE} that is jointly trained with the reconstruction and generative adversarial error losses. In contrast to the LSA, it tries to force the encoder output distribution to be approximately uniform. This causes the decoder to reconstruct just normal outputs for normal and anomalous inputs, resulting in a higher mean squared error for abnormal input datum. 

MemAE \cite{gong2019memorizing} is the first \ac{AE} framework that uses memory in a non-parametric approach for novelty detection. When an input is passed to the \ac{AE}, it is searched within the memory to find embeddings that match the input. Then, based on the combination of these embeddings, a new one is made and passed through the decoder.

Deep SVDD \cite{ruff2018deep} is a one-class classification method that tries to convert data from the original space to the desired space by using deep neural networks. During the training process, it tries to put normal datum in a circle with a predefined center in a new space and then iteratively reduces its diameter. Because of the problem of finding trivial solutions, it uses early stopping and utilizes some constraints on the activation functions of layers. \ac{GT} \cite{golan2018deep} is another one-class classification method that uses self-supervised learning. It makes a set of different transformations on the training data. Using a classifier, it tries to guess which transformation is done on each input to ensure that normal ones are classified correctly at the testing time despite anomalous inputs.

AnoGAN \cite{schlegl2017unsupervised} is the first \ac{GAN} based framework for novelty detection. It trains a \ac{GAN} on the normal training datum. Then at the testing time, it looks for an appropriate latent vector such that the mean generator reconstruction error becomes lower than a certain threshold. Because of its high testing time, the authors in \cite{akcay2018ganomaly} introduced an improved version of AnoGAN, called Ganomaly, which does not need to solve any optimization problem at the testing time. Ganomaly employs \ac{VAE} \cite{kingma2013auto} and \ac{GAN} \cite{goodfellow2014explaining} frameworks to achieve its goal. Rather than solving the optimization problem at the testing time, it just uses the encoder part of \ac{VAE} to obtain the desired latent vector.

% ------------------method--------------------
\section{Method}

We propose the Puzzle-\ac{AE} framework as a mixture of self-supervised learning methods and regular \ac{AE}s to use the good features of both and reduce their important weaknesses.

% As mentioned earlier, self-supervised learning methods are effective in extracting semantically meaningful and generalizable features. On the other hand, the primary deficiency of \ac{AE}s is the inability to generalize normal data on complex datasets under a limited number of training normal samples. For example, in the class car of the CIFAR-10 dataset, \ac{AE} may reconstruct a normal red car but gets confused by a precisely similar input with a different color. Making rich datasets with millions of datum may solve this deficiency, but it is against the premise of novelty detection. Because in many real applications of novelty detection, it is assumed that rich datasets for the normal samples that include every possible variation is not available. Moreover, it is not always feasible to build such a dataset. To use the good features of \ac{AE}s and reduce its important weaknesses, we propose the Puzzle-\ac{AE} framework.

\subsection{Model Training}
The proposed framework is illustrated schematically in Fig. \ref{method}.
U-Net \cite{li2018h} architecture is used as our base framework that has a similar structure to \ac{AE} but is popular because of its ability to reconstruct high-quality images, which is used by many segmentation methods \cite{li2018h,litjens2017survey,wang2017chestx}. However, we pass puzzled input rather than the noisy or original image, and it is expected from the U-Net \cite{li2018h} to reconstruct the right ordered image. The U-Net \cite{li2018h} is trained using the \ac{MSE} loss of its output and original input.  

\textbf{Puzzle Making: } According to the principles of self-supervised learning, we use puzzle-solving as our pretext task. Each input image is split into four partitions, and then a random permutation of these partitions with at least two displacements is selected. The 4-partition puzzle is chosen because it is the most obvious way of making a puzzle, resulted in agnosticism about datasets. To obtain better features, the puzzle-making procedure is combined with inpainting (for gray-scale images) or colorization (for colorful images). Thus, a partition is selected randomly to get entirely black or grayscale accordingly.

Suppose we have a given input $x \in \mathcal{X}$, where $\mathcal{X}$ denotes the entire training dataset. We first split this input into four partitions and convert a random partition to entirely black or grayscale to obtain $\tilde{x}$. We show the set of puzzles as $\mathcal{F} = \left\{f_{P_{i}}(\cdot) \mid i=1, \ldots, K\right\}$ where $K$ is equal to 23 in case of having four partitions in each puzzle and considering all the possible permutations of the partitions with at least two displacements. $f_{P_{i}}(\tilde{x})$ denotes a specific permutation of the four partitions in which one partition is fully blacked or converted to black and white. 

\textbf{Puzzle Making for Texture Images: } The defined set of puzzles $\mathcal{F}$ determines the ambiguity of our self-supervised learning task. As mentioned in \cite{noroozi2016unsupervised}, a good self-supervised learning task should not be ambiguous. Considering texture-like images such as the five texture categories in the MVTecAD dataset, the four partitions in the puzzled image are mostly similar. Consequently, finding the right solution for all the 23 permutations of the partitions would be impossible. Therefore, to make our self-supervised learning task less ambiguous, only the six different permutations of the partitions with precisely two displacements are considered in our puzzle set $\mathcal{F}$ which means $K$ would be equal to 6 in this case. 

\begin{figure}[htb]
\centering
     \includegraphics[width=\linewidth]{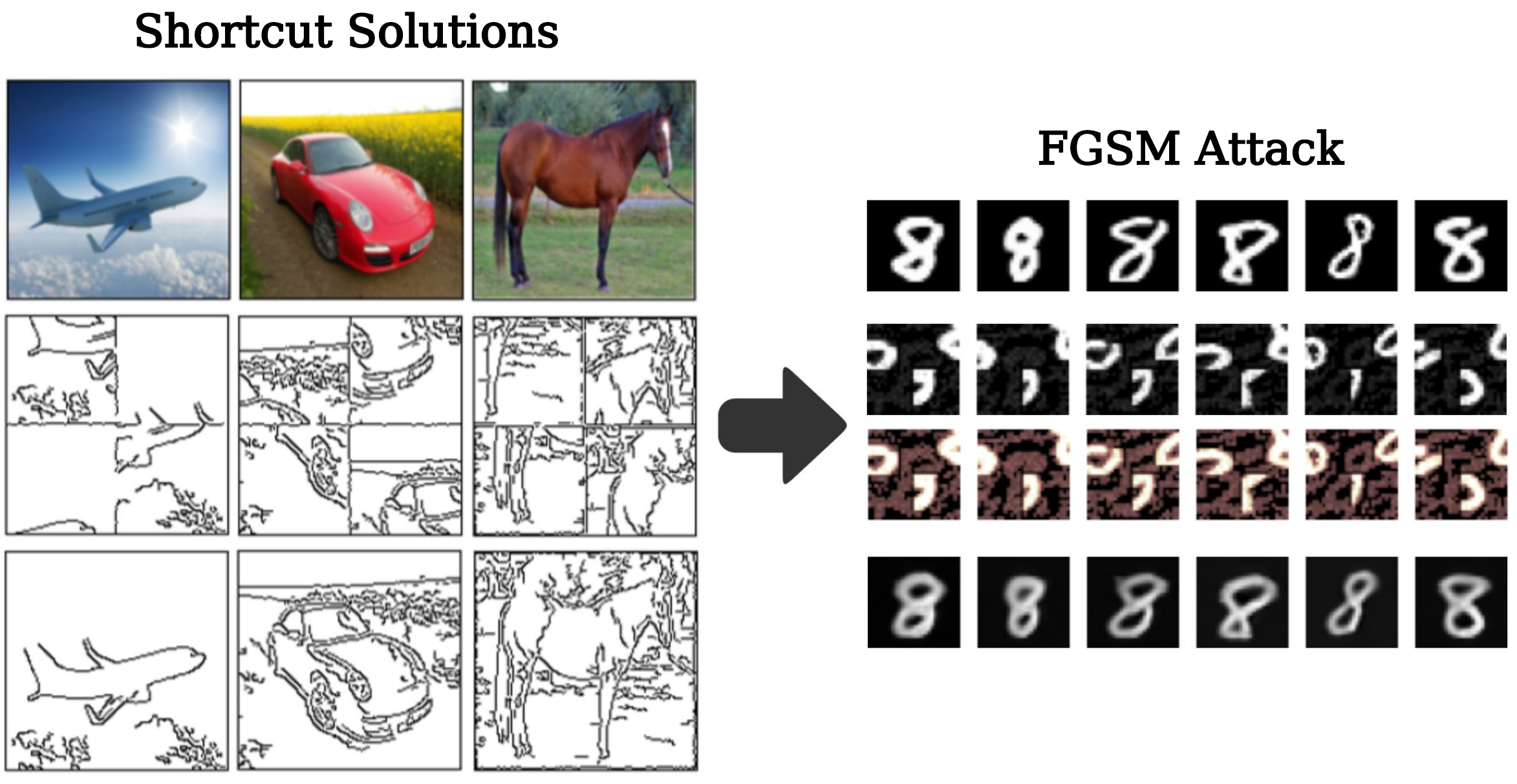}
    %   \caption{As it is shown, by puzzling an image, some trivial features are produced that conduct the model to learn trivial solutions of doing the puzzle. For example, model could understand the number of displacements just by noticing the vertical and horizontal lines. As the Figure shows, FGSM\cite{wong2020fast} tries to make specific anti-shortcut noises that appear similar to number 8. This forces the network not to overfit on low level statistics automatically. Heat map on the third row shows the normal image with better noise clarification.}
    \caption{Some trivial features are produced that conduct the model to learn shortcuts. For example, model could understand the number of displacements just by noticing the vertical and horizontal lines. FGSM \cite{wong2020fast} makes specific anti-shortcut noises that appear similar to the number 8 (right figure). Heat map on the third row shows the normal image with better noise clarification.}
      
      \label{edge_fgsm}
\end{figure}

\textbf{Adversarial Robust Training:} To increase robustness and avoid trivially or shortcut solutions in self-supervised learning methods such as finding low-level statistics of different partitions and finding partitions' border edges in the puzzled input. We obtain robust adversarial training as in \cite{salehi2020arae} (that can be viewed as automatic shortcut removal) to generate adversarial examples. We should mention that to generate these adversarial samples, all we need is a differentiable function. Also, all of the inputs in both training and testing time have at least two displacements in their partitions. Since PGD \cite{madry2017towards} has a high time complexity, we use FGSM \cite{wong2020fast}, instead. Fig. \ref{edge_fgsm} illustrates the problem above and the effect of robust adversarial training on solving it. As it is shown, FGSM \cite{wong2020fast} noise is added to the image purposefully to relieve the effect of low-level statistics such as edges. Eq. \ref{equation:fgsm} shows more details about the manipulation of FGSM \cite{wong2020fast} in the proposed framework. 

\begin{equation}
\label{equation:fgsm}
\left\{\begin{array}{l}
1. \; \delta=Uniform(-\epsilon, \epsilon) \\ 
2. \; \delta=\delta+\alpha \cdot \operatorname{sign}\left(\nabla_{x}\left\|U\left(f_{P_{i}}(\tilde{x})+\delta\right)-f_{P_{i}}(\tilde{x})\right\|_{2}\right) \\ 
3. \; \delta=\max (\min (\delta, \epsilon),-\epsilon) \\
4. \; \tilde{f}_{P_{i}}(\tilde{x})=f_{P_{i}}(\tilde{x})+\delta , \end{array}\right.
\end{equation}
where $U(\cdot)$ indicates the U-Net network, $\epsilon$ is the attack magnitude, and $\alpha$ is the step size. $\tilde{f}_{P_{i}}(\tilde{x})$ is the adversarial sample obtained from the puzzled input $f_{P_{i}}(\tilde{x})$. 
% There exist trivial solutions in self-supervised learning methods that are reputed as shortcut solutions in the literature. There also exist shortcut solutions such as finding low-level statistics of different partitions and using them to do the puzzle, finding partitions border edges in the puzzled input to extract the relational position of partitions, etc. To relieve these shortcut problems, we use robust adversarial training as in \cite{salehi2020arae} (that can be viewed as automatic shortcut removal), and all of the inputs in both training and testing time have at least two displacements in their partitions. Since PGD\cite{madry2017towards} has a high time complexity, we use FGSM\cite{wong2020fast}, instead. 

\begin{figure*}
\centering
     \includegraphics[width=\textwidth]{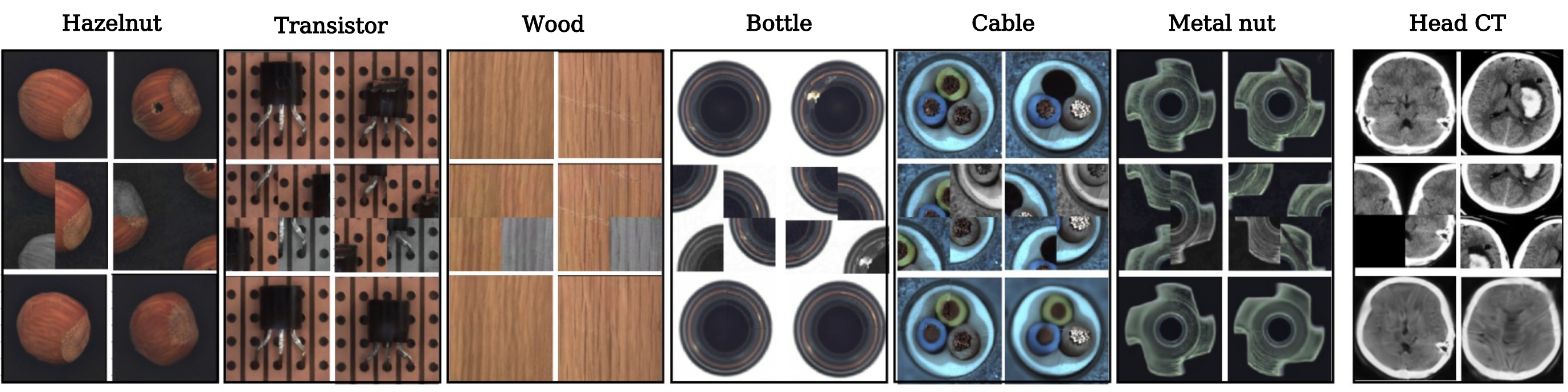}
      \caption{Visualization of the proposed method on MVTecAD \cite{bergmann2019mvtec} and Head CT datasets. First row is the original image, second row is the puzzled input mingled by colorization or inpainting and third row is the model's output. For each category, first column is a normal input and the second column is an anomalous one.}
      \label{MVTec_method}
\end{figure*}

\textbf{Adversarial Training and Total Loss:} The whole framework is also trained similar to the \ac{GAN} framework \cite{goodfellow2014explaining} to improve the quality of the produced images. Better quality is obtained because of the ability of the adversarial training to converge to one mode despite \ac{MSE} that converges to the average of different modes \cite{sabokrou2018avid,chen2018unsupervised,tolstikhin2017wasserstein}. 
We define the reconstruction loss $\mathcal{L}_{rec}$ as the $\mathcal{L}_{2}$ distance between the original input $x$ which is drawn from the input data distribution $p_x$, and the solved puzzle at the output of the U-Net network $U(\tilde{f}_{P_{i}}(\tilde{x}))$:
\begin{equation}
\mathcal{L}_{rec}=\mathbb{E}_{x \sim p_{x}}\left\|U\left(\tilde{f}_{P_{i}}(\tilde{x})\right)-x\right\|_{2}.
\end{equation}
Similar to \cite{akcay2018ganomaly}, feature matching loss in the adversarial training is used. $f_D(\cdot)$ denotes a function representing an intermediate layer of our discriminator $D$. The adversarial loss $\mathcal{L}_{adv}$ is defined as follows:
\begin{equation}
\mathcal{L}_{adv}=\mathbb{E}_{x \sim p_{x}}\left\|f_{D}(x)-\mathbb{E}_{x \sim P_{x}} f_{D}\left(U\left(\tilde{f}_{P_{i}}(\tilde{x})\right)\right)\right\|_{2} .
\end{equation}
Finally, the total loss used as our total training objective is defined as:
\begin{equation}
\mathcal{L}_{total} = \mathcal{L}_{rec} + \lambda \mathcal{L}_{adv},
\end{equation}
where $\lambda$ is a hyper-parameter defining the weight of the $L_{adv}$ term in the total loss.
\subsection{Anomaly Score} The same puzzle-making procedure is used for evaluation. We split each test data into 4 partitions and consider all the $K$ permutations of these partitions with at least 2 displacements without any inpainting or colorization auxiliary task. Suppose $x$ is a given test data and the $i^{\text{th}}$ permutation is used for the puzzle-making process to obtain $f_{P_{i}}(x)$. The anomaly score for this specific permutation is defined as:
\begin{equation}
\mathcal{S}^{\: i}_{test} = \|U(f_{P_{i}}(x)) - x \|_{2} .    
\end{equation} 

\textbf{Error Normalization: } Since solving each of the $K$ puzzles induce different difficulties for the model, the reconstruction error for some puzzles can be much larger than the others. Hence, different reconstruction errors can be obtained for a single input depending on the permutation used for the puzzle-making process. Therefore, we use validation data to normalize these errors over all the permutations. For this purpose, the average reconstruction error over all the validation data is computed for every single permutation. During test time, each reconstruction error is divided by the average error of validation data corresponding to that permutation :
\begin{equation}
\mathcal{S}_{test}^{\: \prime \; i}(x) = \frac{\|U(f_{P_{i}}(x)) - x \|_{2}}{\mathbb{E}_{x \sim p_{x}} \|U(f_{P_{i}}(x)) - x \|_{2}} .
\end{equation}
Finally, min, max and average anomaly score for all the $K$ permutations of a single test data is computed:
\begin{equation}
\mathcal{S}_{test}(x)=\{\min \; \text{OR} \; \max \; \text{OR} \; avg \} _{1\leq i \leq K}\left\{\mathcal{S}_{test}^{\: \prime \; i}(x)\right\} .
\end{equation}
The experiments show that using the max anomaly score is better for simple datasets, and as the dataset becomes more complex, using the average or min would yield better results. Therefore, to avoid unnecessary confusion and complexity in our performance measurement, we report the results by taking the max for the toy datasets and taking the average for the real-world datasets. 

Fig. \ref{MVTec_method} shows the output of the model for some normal and anomalous inputs from MVTecAD \cite{bergmann2019mvtec} and Head CT (hemorrhage) dataset.

% First table: mnist, fashionmnist, cifar10
\begin{table*}[!t]
\centering
\caption{AUROC in \% for several datasets. As it is shown, Puzzle-\ac{AE} surpass the \ac{SOTA} by 7\% on average on the CIFAR-10 \cite{cifar} dataset while it is still competitive on the other ones.}
\label{table:mnist,fashionmnist,CIFAR-10}
\resizebox{\textwidth}{!}{\begin{tabular}{ccccccccccccc}
\hline\noalign{\smallskip}
Dataset & Method & 0 & 1 & 2 & 3 & 4 & 5 & 6 & 7 & 8 & 9 & Mean\\
\noalign{\smallskip}
\hline
\noalign{\smallskip}
% capsnet: 1911.08616
% rest: ARAE
\multirow{7}{*}{MNIST \cite{lecun2010mnist}} & ARAE \cite{salehi2020arae} & 99.8 & 99.9 & 96.0 & 97.2 & 97.0 & 97.4 & 99.5 & 96.9 & 92.4 & 98.5 & 97.5\\
& OCSVM \cite{chen2001one} & 99.5 & 99.9 & 92.6 & 93.6 & 96.7 & 95.5 & 98.7 & 96.6 & 90.3 & 96.2 & 96.0\\
& AnoGAN \cite{schlegl2017unsupervised} & 96.6 & 99.2 & 85.0 & 88.7 & 89.4 & 88.3 & 94.7 & 93.5 & 84.9 & 92.4 & 91.3\\
& DSVDD \cite{ruff2018deep} & 98.0 & 99.7 & 91.7 & 91.9 & 94.9 & 88.5 & 98.3 & 94.6 & 93.9 & 96.5 & 94.8\\
& CapsNet\textsubscript{PP} \cite{li2019icml} & 99.8 & 99.0 & 98.4 & 97.6 & 93.5 & 97.0 & 94.2 & 98.7 & 99.3 & 99.0 & 97.7\\
& OCGAN \cite{perera2019ocgan} & 99.8 & 99.9 & 94.2 & 96.3 & 97.5 & 98.0 & 99.1 & 98.1 & 93.9 & 98.1 & 97.5\\
& LSA \cite{abati2019latent} & 99.3 & 99.9 & 95.9 & 96.6 & 95.6 & 96.4 & 99.4 & 98.0 & 95.3 & 98.1 & 97.5\\
\noalign{\smallskip}
\hline
\noalign{\smallskip}
% max: ganomaly_gt_mnist-fmnist
& OURS & ${99.6\pm0.004}$ & ${99.93\pm0.004}$ & ${97.12\pm0.083}$ & ${96.97\pm0.039}$ & ${97.70\pm0.017}$ & ${98.43\pm0.022}$ & ${99.29\pm0.041}$ & ${98.26\pm0.036}$ & ${94.14\pm0.073}$ & ${98.57\pm0.082}$ & \textbf{98.00}\\
\noalign{\smallskip}
\hline
\noalign{\smallskip}

% ARAE
% DSVDD: pic
\multirow{6}{*}{Fashion-MNIST \cite{xiao2017fashion}} & ARAE \cite{salehi2020arae} & 93.7 & 99.1 & 91.1 & 94.4 & 92.3 & 91.4 & 83.6 & 98.9 & 93.9 & 97.9 & \textbf{93.6}\\
& OCSVM \cite{chen2001one} & 91.9 & 99.0 & 89.4 & 94.2 & 90.7 & 91.8 & 83.4 & 98.8 & 90.3 & 98.2 & 92.8\\
& DAGMM \cite{zong2018deep} & 30.3 & 31.1 & 47.5 & 48.1 & 49.9 & 41.3 & 42.0 & 37.4 & 51.8 & 37.8 & 41.7\\
& DSEBM \cite{zhai2016deep} & 89.1 & 56.0 & 86.1 & 90.3 & 88.4 & 85.9 & 78.2 & 98.1 & 86.5 & 96.7 & 85.5\\
& DSVDD \cite{ruff2018deep} & 98.2 & 90.3 & 90.7 & 94.2 & 89.4 & 91.8 & 83.4 & 98.8 & 91.9 & 99.0 & 92.8\\
& LSA \cite{abati2019latent} & 91.6 & 98.3 & 87.8 & 92.3 & 89.7 & 90.7 & 84.1 & 97.7 & 91.0 & 98.4 & 92.2\\
\noalign{\smallskip}
\hline
\noalign{\smallskip}
% max: Puzzle_AE_Fmnist_Ganomaly 
& OURS(4-parts) & ${91.37\pm0.200}$ & ${98.96\pm0.019}$ & ${89.34\pm0.056}$ & ${92.04\pm0.317}$ & ${91.04\pm0.087}$ & ${90.73\pm0.189}$ & ${82.39\pm0.077}$ & ${98.23\pm0.026}$ & ${91.02\pm0.190}$ & ${97.52\pm0.200}$ & 92.26\\

& OURS(9-parts)\footnotemark[2] & ${91.73\pm0.621}$ & ${98.74\pm0.133}$ & ${89.92\pm0.252}$ & ${91.94\pm0.612}$ & ${89.69\pm0.566}$ & ${93.54\pm0.606}$ & ${84.90\pm0.405}$ & ${98.78\pm0.078}$ & ${92.30\pm1.178}$ & ${98.46\pm0.169}$ & 93.00\\

\noalign{\smallskip}
\hline
\noalign{\smallskip}

% ARAE: Flock 
% lsa, capsnet: 1911.08616
% reset: OCGAN paper
\multirow{7}{*}{CIFAR-10 \cite{cifar}} & ARAE \cite{salehi2020arae} & 72.2 & 43.1 & 69.0 & 55.0 & 75.2 & 54.7 & 70.1 & 51.0 & 72.2 & 40.0 & 60.23\\
% from ocgan paper
& OCSVM \cite{chen2001one} & 63.0 & 44.0 & 64.9 & 48.7 & 73.5 & 50.0 & 72.5 & 53.3 & 64.9 & 50.8 & 58.56\\
& AnoGAN \cite{schlegl2017unsupervised} & 67.1 & 54.7 & 52.9 & 54.5 & 65.1 & 60.3 & 58.5 & 62.5 & 75.8 & 66.5 & 61.79\\
& DSVDD \cite{ruff2018deep} & 61.7 & 65.9 & 50.8 & 59.1 & 60.9 & 65.7 & 67.7 & 67.3 & 75.9 & 73.1 & 64.81\\
& CapsNet\textsubscript{PP} \cite{li2019icml} & 62.2 & 45.5 & 67.1 & 67.5 & 68.3 & 63.5 & 72.7 & 67.3 & 71.0 & 46.6 & 61.2\\
& OCGAN \cite{perera2019ocgan} & 75.7 & 53.1 & 64.0 & 62.0 & 72.3 & 62.0 & 72.3 & 57.5 & 82.0 & 55.4 & 65.66\\
& LSA \cite{abati2019latent} & 73.5 & 58.0 & 69.0 & 54.2 & 76.1 & 54.6 & 75.1 & 53.5 & 71.7 & 54.8 & 64.1\\
\noalign{\smallskip}
\hline
\noalign{\smallskip}
% avg: lsa-gt-ocgan-coil
& OURS & ${78.93\pm0.203}$ & ${78.05\pm0.755}$ & ${69.95\pm0.344}$ & ${54.88\pm0.410}$ & ${75.46\pm0.204}$ & ${66.04\pm0.430}$ & ${74.76\pm0.280}$ & ${73.30\pm0.468}$ & ${83.34\pm0.256}$ & ${69.96\pm0.461}$ & \textbf{72.47}\\
\noalign{\smallskip}
\hline
% \noalign{\smallskip}

\end{tabular}}
\end{table*}

% ---------------------------Experiments------------------
\section{Experiments}
In this section, we validate our method by conducting extensive experiments. We consider multiple commonly used toy and also real-world datasets for evaluating our model. \footnote{The code to reproduce the results is provided at \url{https://github.com/Niousha12/Puzzle_Anomaly_Detection}.}

% In this section, we validate our method by conducting extensive experiments. We consider multiple commonly used datasets such as MNIST, Fashion-MNIST, CIFAR-10, and COIL-100, and also real-world datasets like MVTecAD and medical datasets for evaluating our model. \footnote{Code to reproduce the results is provided at \url{https://github.com/Niousha12/Puzzle_Anomaly_Detection}.} The experiments show that the proposed method outperforms several state-of-the-art approaches on these datasets. \footnote{Code to reproduce the results is provided at \url{https://github.com/Niousha12/Puzzle_Anomaly_Detection}.}

\begin{figure}[!ht]
     \includegraphics[width=\linewidth]{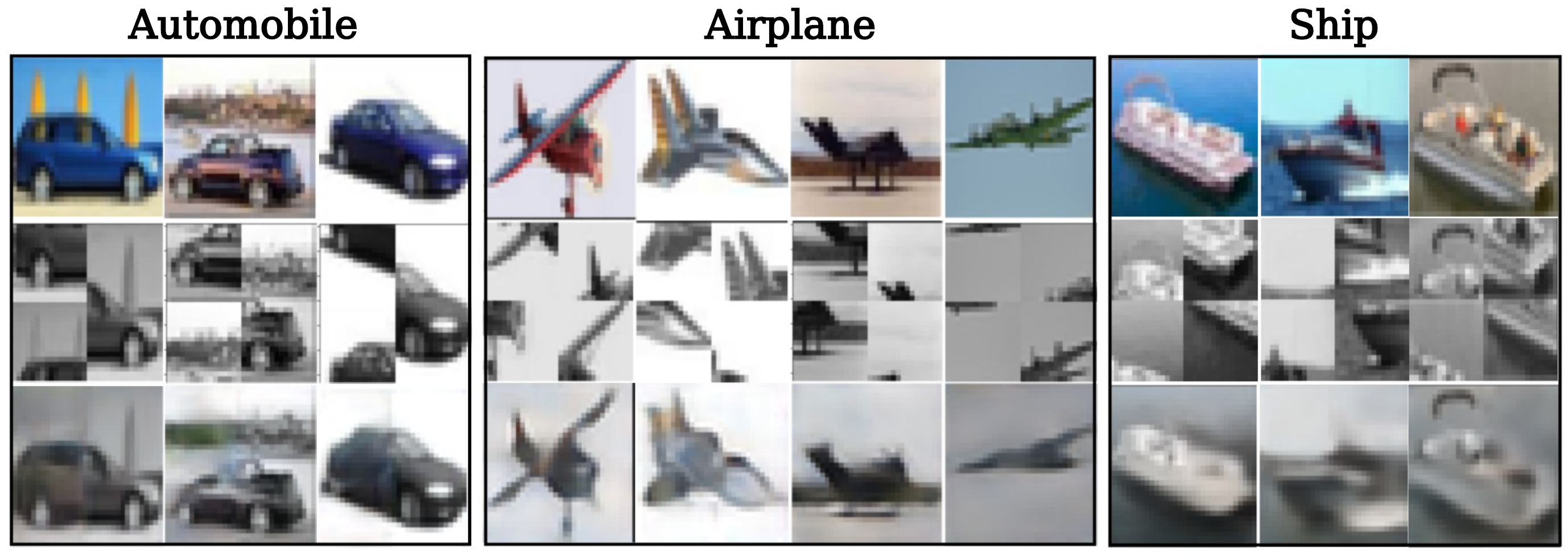}
      \caption{Effect of converting images to grayscale on the model learned features for some classes of CIFAR-10 \cite{cifar} dataset. As it is shown, the model can produce perfect outputs even for grayscale inputs.}
      \label{grayscale}
\end{figure}
% \input{Figures/Medical_method}
% mvtec table
\begin{table*}[!t]
\centering
% \caption{Area under the \ac{ROC} curve in \% on MVTec AD\cite{bergmann2019mvtec} dataset. As it is shown, Puzzle-\ac{AE} surpass other \ac{SOTA}s significantly and about 4.6\% with respect to the best previous \ac{SOTA}.}
\caption{AUROC in \% on MVTec AD \cite{bergmann2019mvtec} dataset.  We surpass the SOTA by $\sim4.6\%$.}
\label{table:MVTec}
\resizebox{\textwidth}{!}{\begin{tabular}{ccccccccccccccccc}
\hline\noalign{\smallskip} 
% Dataset & 
Method & Bottle & Hazelnut & Capsule & Metal Nut & Leather & Pill & Wood & Carpet & Tile & Grid & Cable & Transistor & Toothbrush & Screw & Zipper & Mean\\
\noalign{\smallskip}
\hline
\noalign{\smallskip}
% 1911.08616
AVID \cite{sabokrou2018avid} & 88.0 & 86.0 & 85.0 & 63.0 & 58.0 & 86.0 & 83.0 & 70.0 & 66.0 & 59.0 & 64.0 & 58.0 & 73.0 & 66.0 & 84.0 & 73.0\\
AE\textsubscript{SSIM} \cite{bergmann2019visigrapp} & 88.0 & 54.0 & 61.0 & 54.0 & 46.0 & 60.0 & 83.0 & 67.0 & 52.0 & 69.0 & 61.0 & 52.0 & 74.0 & 51.0 & 80.0 & 63.0\\
AE\textsubscript{L2} \cite{bergmann2019visigrapp} & 80.0 & 88.0 & 62.0 & 73.0 & 44.0 & 62.0 & 74.0 & 50.0 & 77.0 & 78.0 & 56.0 & 71.0 & 98.0 & 69.0 & 80.0 & 71.0\\
AnoGAN \cite{schlegl2017unsupervised} & 69.0 & 50.0 & 58.0 & 50.0 & 52.0 & 62.0 & 68.0 & 49.0 & 51.0 & 51.0 & 53.0 & 67.0 & 57.0 & 35.0 & 59.0 & 55.0\\
LSA \cite{abati2019latent} & 86.0 & 80.0 & 71.0 & 67.0 & 70.0 & 85.0 & 75.0 & 74.0 & 70.0 & 54.0 & 61.0 & 50.0 & 89.0 & 75.0 & 88.0 & 73.0\\
\noalign{\smallskip}
\hline
\noalign{\smallskip}
% avg from two different table: Puzzle_AE_MVTec_Augmented800_Ganomaly
OURS & ${94.24\pm0.10}$ & ${91.21\pm0.13}$ & ${66.88\pm0.23}$ & ${66.33\pm0.10}$ & ${72.86\pm0.86}$ & ${71.63\pm0.11}$ & ${89.51\pm0.63}$ & ${65.73\pm0.37}$ & ${65.48\pm0.12}$ & ${75.35\pm0.60}$ & ${87.90\pm0.10}$ & ${85.96\pm0.12}$ & ${97.79\pm0.05}$ & ${57.81\pm1.12}$ & ${75.74\pm0.13}$ & \textbf{77.63}\\
\noalign{\smallskip}
\hline
% \noalign{\smallskip}

\end{tabular}}
\end{table*}

\subsection{Experimental Setup}
\textbf{Datasets.} We considered seven datasets for evaluating our method: MNIST, Fashion-MNIST, CIFAR-10, COIL-100, MVTecAD, and two medical datasets (Head CT (hemorrhage) and Brain MRI Images for Brain Tumor Detection). We briefly describe each of these datasets:
\\
\textbf{MNIST \cite{lecun2010mnist}}: 60k training and 10k test $28 \times 28$ gray-scale handwritten digit images from 0 to 9. 
\textbf{Fashion-MNIST \cite{xiao2017fashion}}: 60k training and 10k test grayscale images of 10 fashion product categories. 
\textbf{CIFAR-10 \cite{cifar}}: 50k training and 10k test $32 \times 32$ color images in 10 classes.
\textbf{COIL-100 \cite{nene1996columbia}}: 7200 $128 \times 128$ color images of 100 object classes with 72 images of each object in different poses.
\textbf{MVTecAD \cite{bergmann2019mvtec}}: an industrial dataset with more than 5k high-resolution images in 15 categories of objects and textures. Each category contains both normal and anomalous images with various kinds of defects (used for testing). We downscale all images to the size $128 \times 128$ and use zoom data augmentation to create 800 training images for each class. 
\textbf{Head CT (hemorrhage)} \cite{kitamura2018hemorrhage}: a medical dataset containing 100 $128 \times 128$ normal head CT images and 100 with hemorrhage.
\textbf{Brain MRI Images for Brain Tumor Detection} \cite{chakrabarty2019tumor}: a medical dataset with 98 $256 \times 256$ normal MRI images and 155 with brain tumors. \\
% \end{itemize}

\footnotetext[2]{The whole procedure is entirely the same as the 4-part puzzle. However, the datum is extended to $30 \times 30$ and partitioned into nine parts where six parts are permuted, and one is fully blacked randomly.}

\textbf{Model Configuration and Hyperparameters.} We used the standard U-Net architecture without the batchnorm layers \cite{ioffe2015batch} as our base framework, which is trained to reconstruct right order images. Moreover, the common discriminator introduced in DCGAN \cite{radford2015unsupervised} was used as our next subnetwork, which was trained to classify the original input and output of the U-Net \cite{li2018h} as real or fake. We used the Adam optimizer \cite{kingma2014adam} for training both networks. The learning rate was initially set to 1e-3 for the U-Net \cite{li2018h} and 2e-4 for the discriminator. We used a learning rate scheduler to multiply the learning rate by 0.8 when the minimum amount of loss did not change for 50 subsequent epochs \cite{goyal2017accurate}. Finally, we used the FGSM attack for robust adversarial training of the model. 
We trained the model until convergence of the loss function with a batch size of 8 for MVTecAD \cite{bergmann2019mvtec} and medical datasets and a batch size of 128 for the rest of the datasets.

\textbf{Evaluating Protocols.} The data partitioning used for the training-testing procedures is done similarly to \cite{perera2019ocgan} that introduces two protocols. We use Protocol 1 for the COIL-100 \cite{Nene96objectimage} dataset that randomly takes one class as the normal data and other classes as an anomaly. It uses 80\% of all normal samples for the training and the rest for the test time normal samples. Test time anomalies are sampled from other classes until normal data, and anomalous ones are the same. This process is repeated 30 times, and the results are averaged. We randomly selected ten normal images for the medical datasets and used them along with the anomalous ones for the test data. The rest of the normal images were used for the training. For the MVTecAD \cite{bergmann2019mvtec} dataset, we use the given train and test sets for each class. Zoom augmentation is also performed to create 800 training images for each category of MVTecAD. Protocol 2 is used for all other datasets, which uses the whole training set of just one class as the normal data for training and the whole test set for the test time. We consider 15\% of the training data as validation in each dataset. We evaluated the performance using the \ac{AUC} of the \ac{ROC} curve, which is commonly used for measuring performance in anomaly detection tasks. 

\subsection{Training and Testing Computational Cost}
Because our method uses adversarial samples in its training process, one extra cost is added to our normal training process. To keep the additional cost to a minimum level, an FGSM attack has been used, which adds only one backpropagation to the training process. For the testing time, we compare our method with one of the best performing \ac{SOTA}. To compare the execution time of our method with GT \cite{golan2018deep}, we ran the CIFAR-10 experiment on the NVIDIA-GTX1080ti processor with 11 Gigabytes of RAM and with the same batch size. It has been observed that our testing routine has $4.7 \times$ better execution performance than the GT algorithm, and it can become even faster by employing parallelism for finding each of the permutations' costs. This faster execution time is beneficial in time-dependant anomaly detection tasks \cite{chalapathy2019deep}.

% coil table
\begin{table}[htb]
\centering
% \caption{Area under the \ac{ROC} curve in \% on COIL-100\cite{Nene96objectimage} dataset. As it is shown, Puzzle-AE reaches  one of the \ac{SOTA}s and surpass the others signifantly.}
\caption{AUROC in \% on COIL-100 \cite{Nene96objectimage} dataset. Obviously, Puzzle-AE reaches one of the \ac{SOTA}s.}
\label{table:Coil}
{\begin{tabular}{ccccccc}
\hline
% \noalign{\smallskip} 
& COIL-100 \\
\hline
% \noalign{\smallskip} 
% ARAE
% ocgan paper
Outlier Pursuit \cite{xu2010robust} & 90.8 \\
DPCP \cite{tsakiris2018dual} & 90.0 \\
ALOCC DR \cite{sabokrou2018adversarially} & 80.9 \\
ALOCC D \cite{sabokrou2018adversarially} & 68.6 \\
GPND \cite{pidhorskyi2018generative} & 96.8 \\
OCGAN \cite{perera2019ocgan} & \textbf{99.5} \\
% \noalign{\smallskip}

\hline
OURS & \textbf{99.3}\\
\hline

% % max:ours -> lsa-gt-ocgan-coil
% 90.8 & 90.0 & 80.9 & 68.6 & 96.8 & \textbf{99.5} & 99.3 \\
% \noalign{\smallskip}
% \hline
% \noalign{\smallskip}

\end{tabular}}
\end{table}
% Medical table
\begin{table}[htb]
\centering
% \caption{Area under the \ac{ROC} curve in \% on real-world medical datasets. As it is shown, Puzzle-AE has significant improvement with respect to the other \ac{AE} based \ac{SOTA} methods.}
\caption{AUROC in \% on real-world medical datasets. We have significant improvement with respect to the other \ac{AE}-based \ac{SOTA} methods.}
\label{table:Medical}
\resizebox{\linewidth}{!}{\begin{tabular}{cccccccc}
\hline\noalign{\smallskip} 
% lsa-gt-ocgan-coil
& & LSA\textsuperscript{*} \cite{abati2019latent} & OCGAN\textsuperscript{*} \cite{perera2019ocgan} & OURS\\
\noalign{\smallskip}
\hline
\noalign{\smallskip}
\multirow{2}{*}{Head CT}
& AUC & ${81.67\pm0.358}$ & ${51.22\pm3.626}$ & ${\textbf{86.43}\pm\textbf{0.04}}$\\
% \noalign{\smallskip}
% \cline{2-5}
% \noalign{\smallskip}
& FPR & 0.81 & 1.00 & \textbf{0.70}\\
\noalign{\smallskip}
\hline
\noalign{\smallskip}
% ours:avg, with normalization, puzzle_AE_Medical_Ganomaly
\multirow{2}{*}{Brain MRI}
& AUC & ${95.61\pm1.433}$ & ${91.74\pm3.050}$ & ${\textbf{96.34}\pm\textbf{0.031}}$\\
% \noalign{\smallskip}
& FPR & \textbf{0.40} & 0.60 & 0.50\\
\noalign{\smallskip}
\hline

\end{tabular}}
\end{table}

\begin{table}[htb]
\centering
\caption{Significant better generalization of Puzzle-AE in compression with \ac{GT} \cite{golan2018deep}.}
\label{table:GT}
{\begin{tabular}{c c c}
\hline
% \noalign{\smallskip} 
& GT \cite{golan2018deep} & OURS\\
% \noalign{\smallskip}
\hline
% \noalign{\smallskip}

MNIST \cite{lecun2010mnist} & \textbf{98.00} & \textbf{98.00} \\
CIFAR-10 \cite{cifar} & \textbf{82.30} & 72.47 \\
MVTec \cite{bergmann2019mvtec} & $67.06*$ & \textbf{77.63} \\
Head CT & $44.70^*$ & \textbf{86.43} \\
Brain MRI & $82.07^*$ & \textbf{96.34} \\

% \noalign{\smallskip}
\hline

\end{tabular}}
\end{table}
\subsection{Results}

In this section, our method’s results are compared with \ac{SOTA} on the mentioned datasets.
% Tables. \ref{table:mnist,fashionmnist,CIFAR-10}, \ref{table:MVTec}, \ref{table:Coil}, and \ref{table:Medical} show the results. 
We report the mean and variance of our model \ac{AUC} in the last 20 epochs of training. Other methods' results are obtained from the main paper or reproduced from their officially released code.

\begin{figure}[!h]
\centering
     \includegraphics[width=\linewidth]{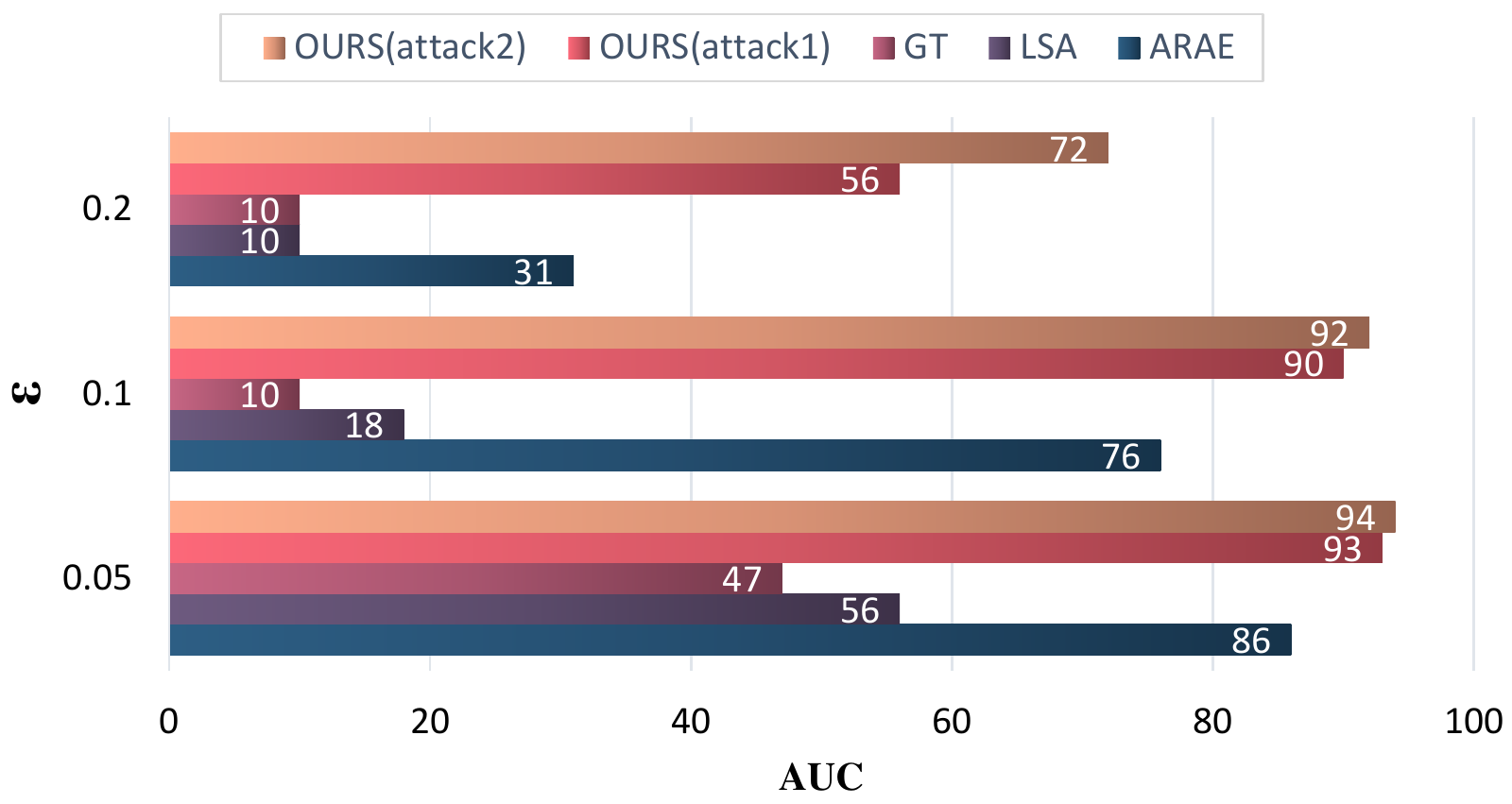}
      \caption{Robustness to adversarial attack on normal data at testing time. The results are shown for three different $\epsilon$ and the model is trained on class 8 of the MNIST \cite{lecun2010mnist} dataset.}
    %   Clearly, Puzzle-AE is significantly more robust than other \ac{SOTA}s on the most challenging class of the MNIST \cite{lecun2010mnist} dataset.}
\label{plot:attack}
\end{figure}

\textbf{AUC comparison with SOTA methods:} The \ac{AUC} results are presented for MNIST \cite{lecun2010mnist}, Fashion-MNIST \cite{xiao2017fashion} and CIFAR-10 \cite{cifar} in Table. \ref{table:mnist,fashionmnist,CIFAR-10}, in which Puzzle-AE is compared with the recent \ac{SOTA}s on these datasets. Table. \ref{table:MVTec} shows comparison between Puzzle-AE and other \ac{SOTA} methods on real-world, industrial dataset MVTecAD \cite{bergmann2019mvtec}. Table. \ref{table:Coil} shows the results of Puzzle-AE on COIL-100 \cite{Nene96objectimage} dataset. To show the effectiveness and generalization properties of the proposed framework, two different experiments are conducted on two different medical datasets, and the results are presented in Table. \ref{table:Medical}. The results verify that Puzzle-AE performs significantly better than the competing methods. We also compared our method with a recent self-supervised learning method called GT \cite{golan2018deep}, and the results are shown in Table. \ref{table:GT}. Moreover, while GT needs extra expert knowledge to design extra-large (72) unambiguous geometric transformations, Puzzle-AE is robust and does not need expert knowledge. Fig. \ref{plot:attack} shows the brittleness of such transformations that become ambiguous with a low amount of noise for competing methods and also shows the robustness of Puzzle-AE in comparison to these methods.

\textbf{AUC comparison with none-AE/AE based SSL methods on MVTecAD:} Although GT \cite{golan2018deep} is an SSL based anomaly detector, we compare our method with other similar tasks to provide a more comprehensive comparison. In this part, the performance of our method is compared with the two of the most well-known SSL methods, such as jigsaw puzzle \cite{noroozi2016unsupervised}, and RotNet \cite{gidaris2018unsupervised}. As it is shown in the Table. \ref{table:Other}, similar to GT, these none AE-based SSL methods substantially fail on the real-world dataset MVTecAD. That is probably because they discard a lot of pixel-level information and only preserve as much as enough to solve their classification tasks. This is desirable, especially when dealing with semantic anomalies such as the CIFAR-10 dataset; however, as the results show, they are weak when dealing with subtle anomalies used in most industrial settings. We also compare puzzle-solving performance with the rotation prediction task implemented on our framework in ablation studies. 
% As the Table. \ref{table:Other} reports, the performance of the rotation task is 1\% lesser than our method on the average of 15 classes. That is because of the rotation invariant aspect existing in most of the texture classes. As mentioned earlier, the puzzle-solving task is less ambiguous for different datasets than rotation prediction, which means we have lesser presumptions on training datasets for various problems. 

% min, max, avg, all in one :))
\begin{table*}[ht]
\centering
\caption{The top rows shows AUROC in \% for some of the fundamental none AE-based SSL methods on MVTecAD. The bottom row shows the AUROC in \% of the Rot-AE on the similar dataset.}
\resizebox{\textwidth}{!}{\begin{tabular}{cccccccccccccccccc}
\hline\noalign{\smallskip} 
% Dataset & 
 & Method & Bottle & Hazelnut & Capsule & Metal Nut & Leather & Pill & Wood & Carpet & Tile & Grid & Cable & Transistor & Toothbrush & Screw & Zipper & Mean\\
\noalign{\smallskip}
\hline
\noalign{\smallskip}
\multirow{2}{*}{None AE-based SSL methods} 
& Jigsaw puzzle \cite{noroozi2016unsupervised} & 38.41 & 57.96 & 41.28 & 52.74 & 12.6 & 47.71 & 46.49 & 35.79 & 24.13 & 60.48 & 33.62 & 24.00 & 35.56 & 51.24 & 8.95 & 37.93\\
& RotNet \cite{gidaris2018unsupervised} & 43.1 & 68.61 & 67.55 & 67.57 & 37.19 & 66.58 & 64.3 & 41.09 & 49.57 & 72.14 & 74.79 & 78.67 & 81.94 & 34.06 & 81 &  61.87\\
\noalign{\smallskip}
\hline
\noalign{\smallskip}
\multirow{1}{*}{AE-based SSL method} 
& Rot-AE & 90.96 & 84.98 & 68.21 & 77.98 & 68.98 & 82.79 & 96.54 & 37.13 & 55.87 & 77.34 & 84.85 & 84.19 & 97.65 & 53.92 & 86.01 & 76.49\\
\noalign{\smallskip}
\hline
\noalign{\smallskip}

\end{tabular}}
\label{table:Other}
\end{table*}

\textbf{FPR comparison for high TPR:} As discussed earlier, the two critical operating points of \ac{ROC} curve are when \ac{TPR} is equal to 99.0\% or 99.5\%. Table \ref{table:FPR_MNIST} shows that Puzzle-AE has significantly lower \ac{FPR} in those critical points in comparison with LSA \cite{abati2019latent}, while their difference in \ac{AUC} is not significant.
\begin{figure}[!t]
\centering
     \includegraphics[width=\linewidth]{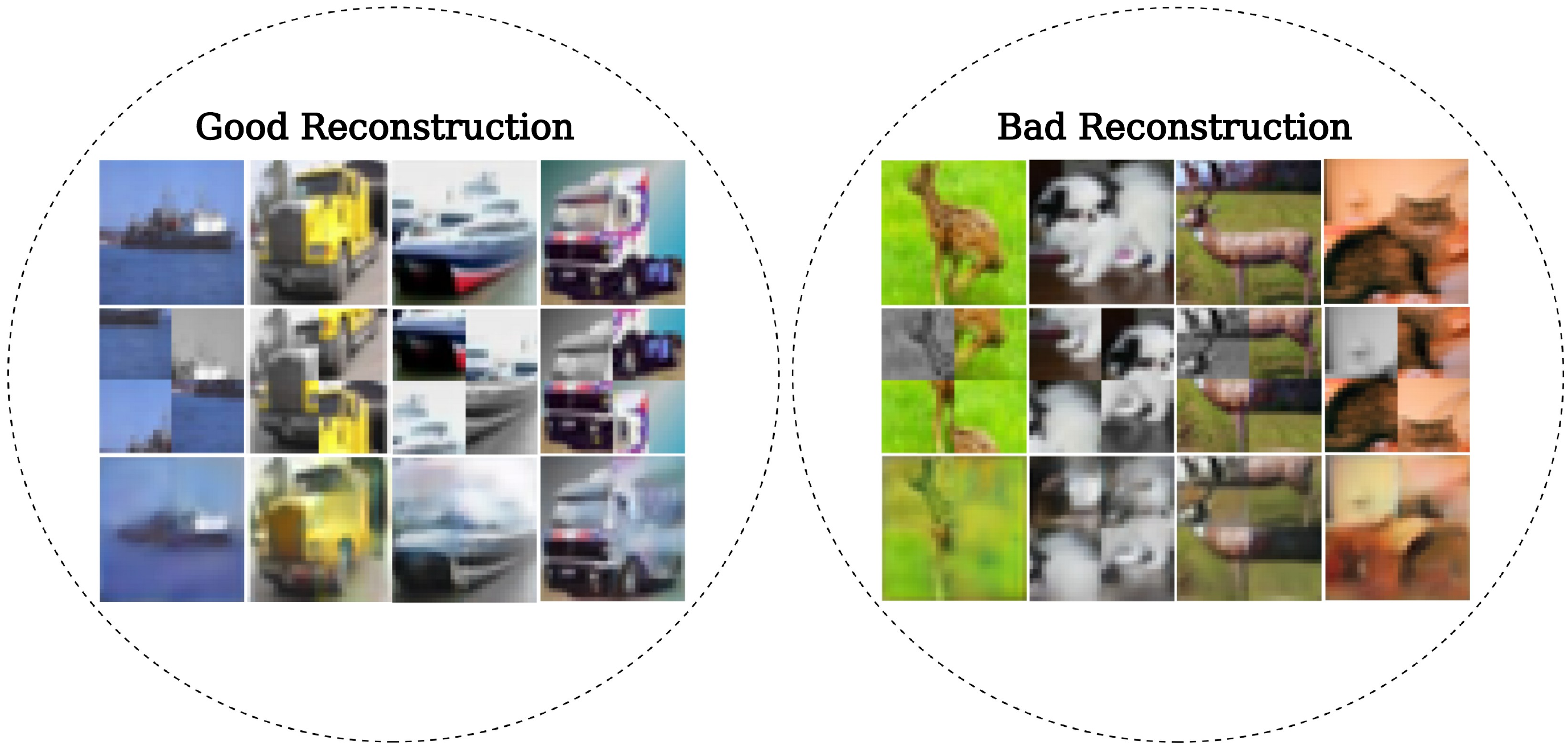}
    %   \caption{Outputs of the model trained on the class car of the CIFAR-10\cite{cifar} dataset for anomalous inputs. Some samples of the truck class are also unpuzzled well, which could be for high similarity between main features of car and truck. The same implications exist for some of the ship class samples.}
    \caption{Anomalous reconstruction of the model trained on the class car of the CIFAR-10 \cite{cifar} dataset. Good reconstruction occurs in classes with high similarity in main features such as truck and ship.}
      \label{AnomalyReconstruction}
\end{figure}
\begin{figure}[!ht]
\centering
     \includegraphics[width=\linewidth]{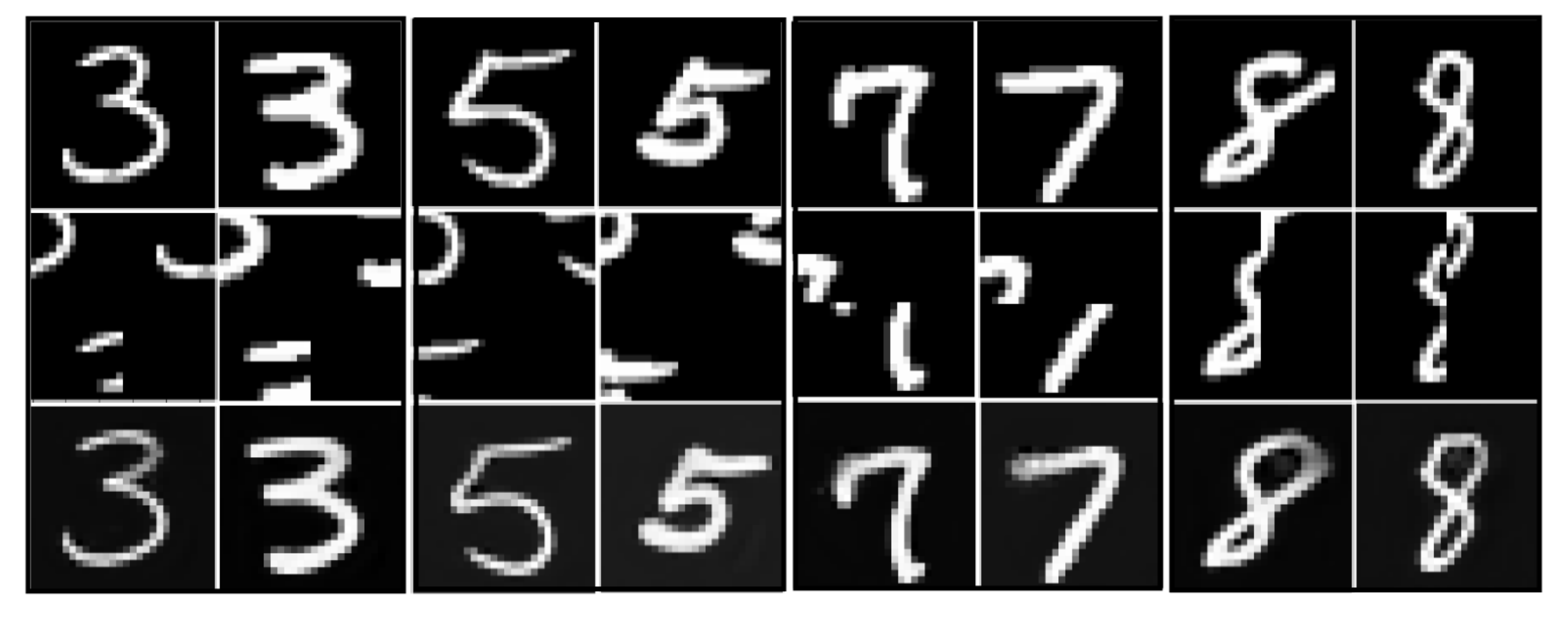}
      \caption{Visualization of the proposed method on MNIST \cite{lecun2010mnist} dataset. First row is the original input, second row is the puzzled input mingled by inpainting task and the third row is the unpuzzled output.}
      \label{Mnist_method}
\end{figure}

\begin{figure}[!thb]
     \includegraphics[width=\linewidth]{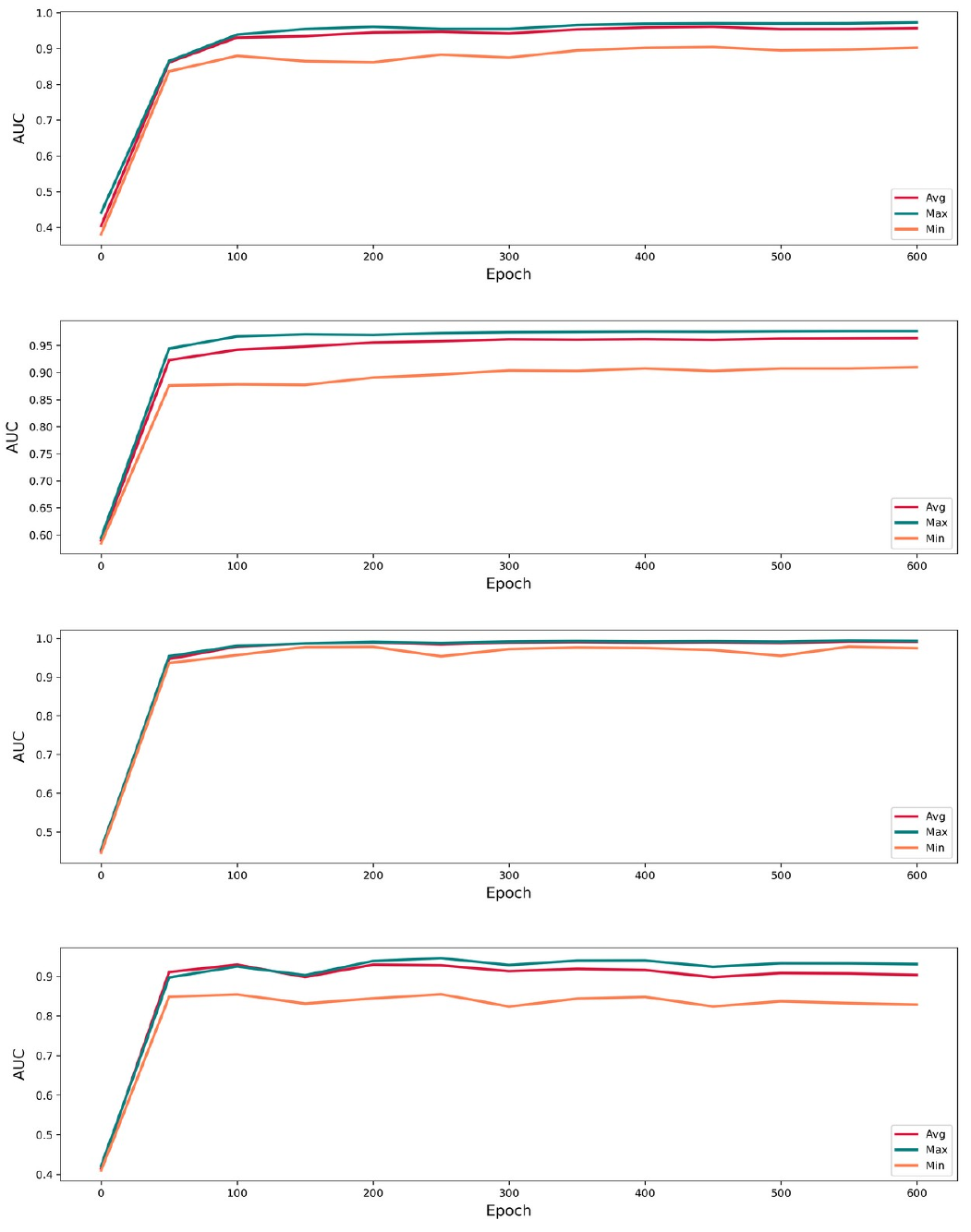}
      \caption{Min, max and avg \ac{AUC} Plots with respect to training epochs are shown for the classes 2, 4, 6 and 8 of the MNIST \cite{lecun2010mnist} dataset.}
      \label{mnist_total}
\end{figure}

\begin{figure}[!thb]
     \includegraphics[width=\linewidth]{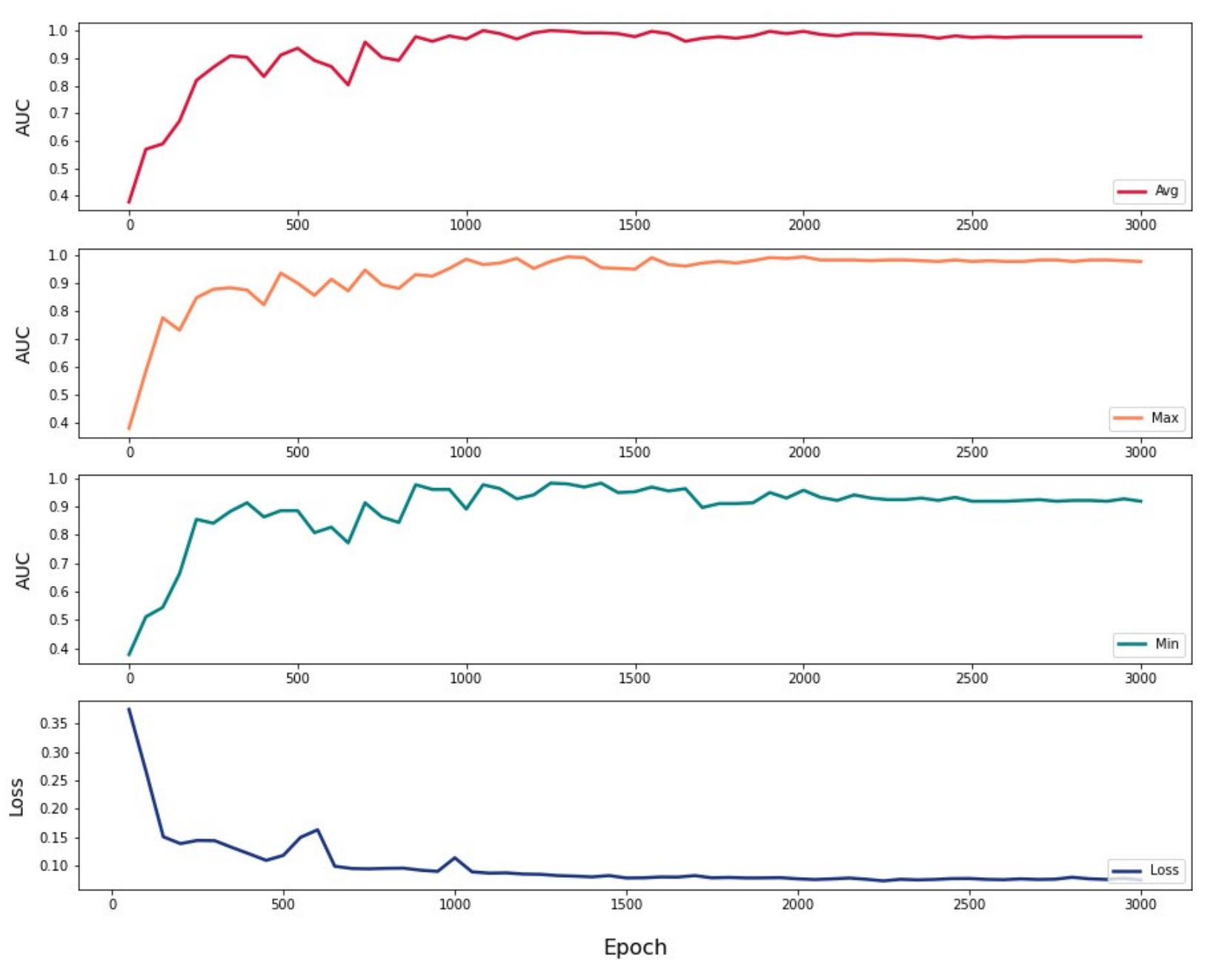}
      \caption{Min, max and avg \ac{AUC} Plots with respect to training epochs are shown for the class toothbrush of the MVTecAD \cite{bergmann2019mvtec} dataset. As it is shown, training procedure continues till the convergence of all of the \ac{AUC} plots and also loss value.}
      \label{Avg_Max_Min_Loss_Plot}
\end{figure}

\begin{figure}[!thb]
     \includegraphics[width=\linewidth]{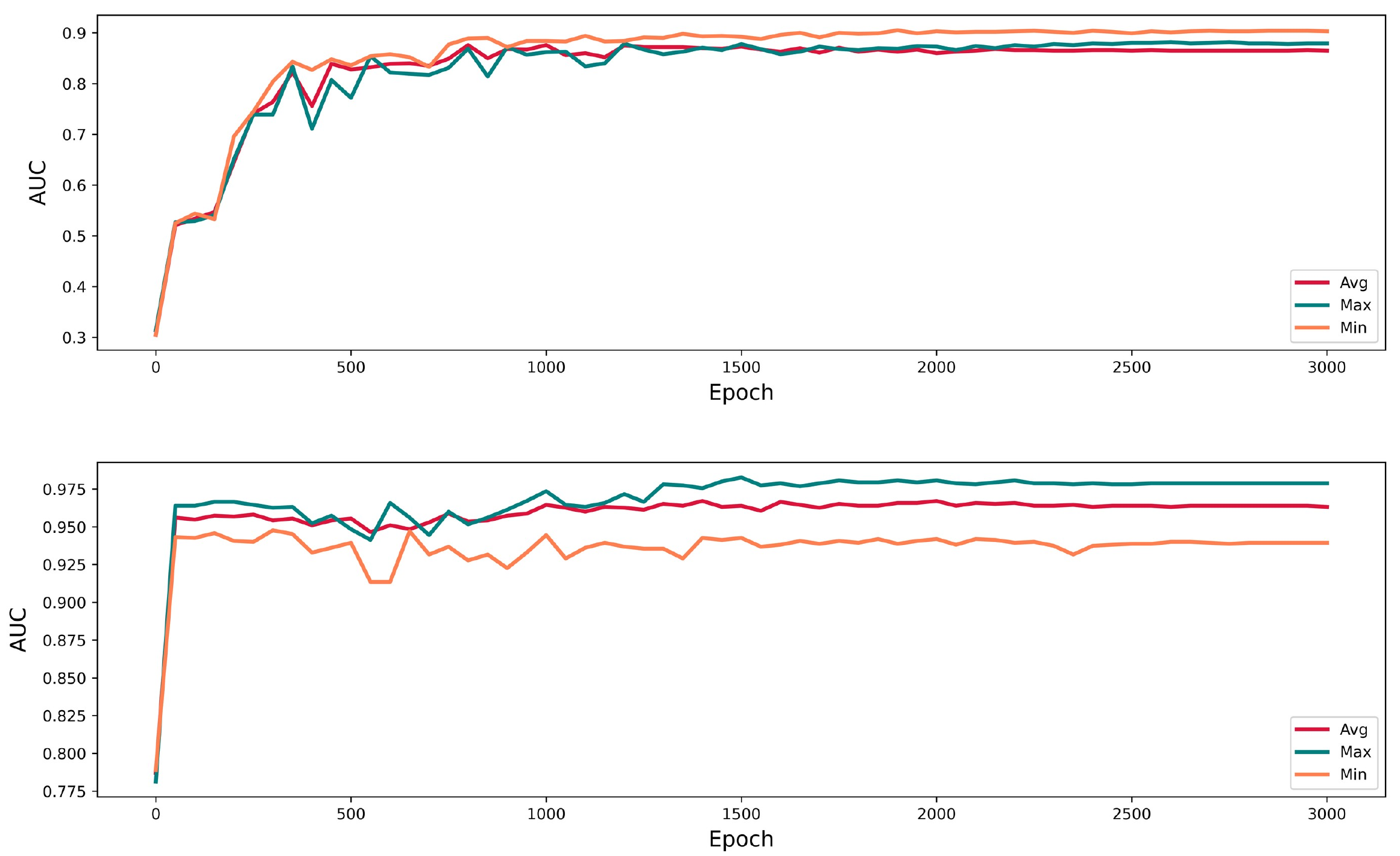}
      \caption{Min, max and avg \ac{AUC} Plots with respect to training epochs are shown for each medical dataset.}
      \label{medical_total}
\end{figure}

\begin{figure}[!thb]
     \includegraphics[width=\linewidth]{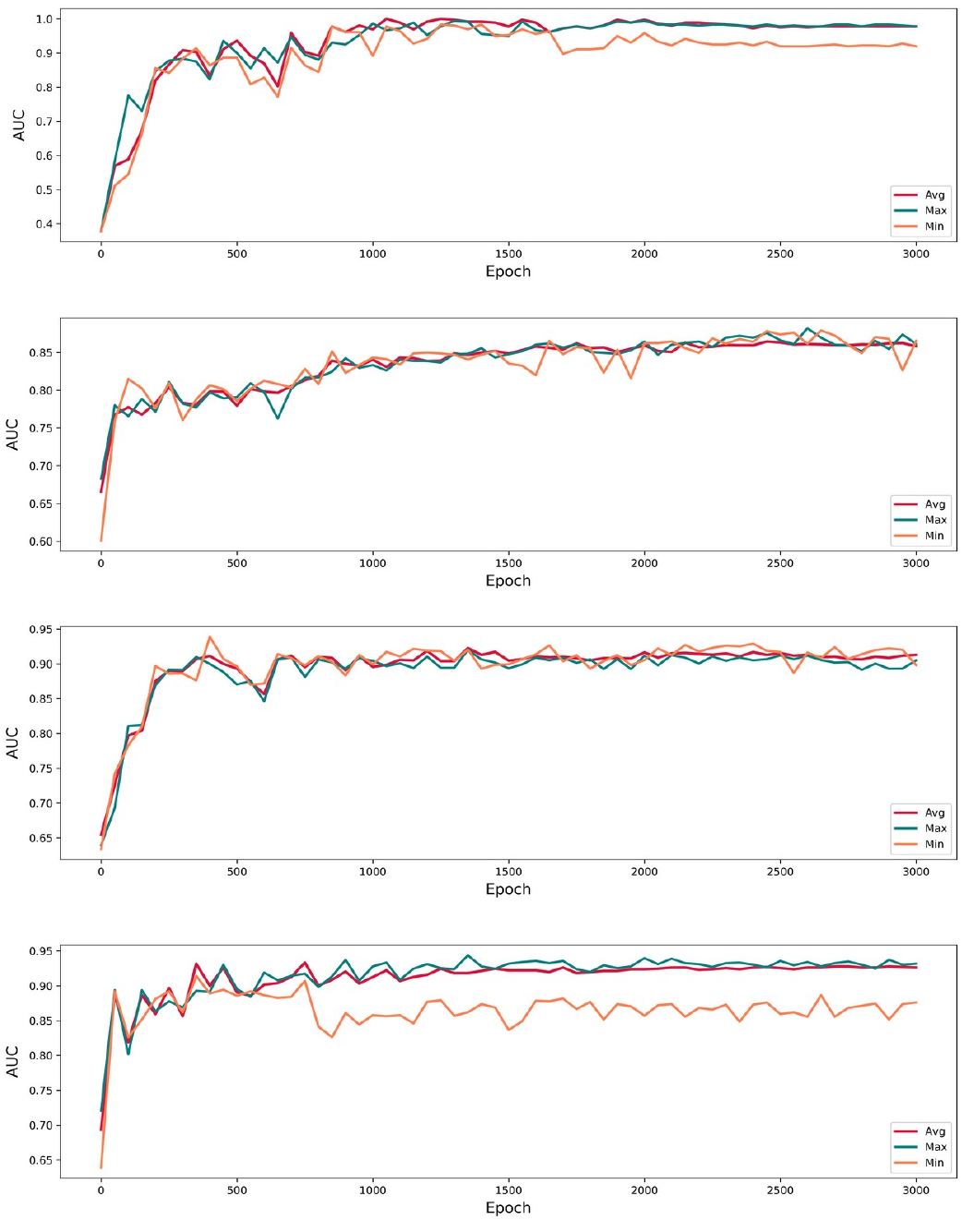}
      \caption{Min, max and avg \ac{AUC} Plots with respect to training epochs are shown for the classes toothbrush, transistor, hazelnut and bottle of the MVTecAD \cite{bergmann2019mvtec} dataset.}
      \label{mvtec_total}
\end{figure}

\textbf{Model stability:} 
Puzzle-AE has more stability in comparison with other methods. Fig. \ref{Avg_Max_Min_Loss_Plot}, \ref{medical_total}, \ref{mvtec_total}, \ref{mnist_total}, and tables \ref{table:mnist,fashionmnist,CIFAR-10}, \ref{table:MVTec}, \ref{table:Medical}  show that we could achieve a highly reliable and stable model with the low variances at the end of the training phase. Moreover, because of using weight decay, the weights of our model are bounded, which is the consequence of Lipschitz continuity of Puzzle-AE. 

Let $w$ denotes the weights of our model and let $\mathcal{L}$ be the total loss. Because of using weight decay, the main objective of the training procedure is defined as:

\begin{equation}
\min_{w} \;\; \mathcal{L} + c\|w\|_{\mathcal{F}}^{2}
\end{equation}
where $\|\cdot\|_{\mathcal{F}}$ denotes the Frobenius norm and $c$ is a constant.
This is equivalent to:
\begin{equation}
\min_{w} \; \mathcal{L} \;\; s.t. \;\; \|w\|_{\mathcal{F}}^{2} \leq M
\end{equation}
which means that the model weights are bounded by some constant $M$. Also because all of the activation functions are smooth so have limited derivative value. This results in having limited  upper bound for the whole network. Furthermore, Puzzle-AE produces smooth output, and it is shown that \cite{chen2018convergence}, being Lipschitz and smooth, the convergence of ADAM optimizer \cite{kingma2014adam} is theoretically provable. This means that we could achieve a highly reliable and stable model at the end of the training phase. This is not the case for other methods such as \ac{GT} \cite{golan2018deep} and DSVDD \cite{ruff2018deep} as shown in the table \ref{table:GT_epoch} where there are large fluctuations of AUC in different epochs of their training process. Because of using unprincipled early stopping methods that usually do not generalize well on unseen datasets, \ac{GT} \cite{golan2018deep} \ac{AUC} fluctuates nearly by 6\% on the medical dataset and nearly 30\% on MVTecAD \cite{bergmann2019mvtec} dataset while its training accuracy is above 98\%.

\begin{figure}[!t]
\centering
     \includegraphics[width=0.95\linewidth]{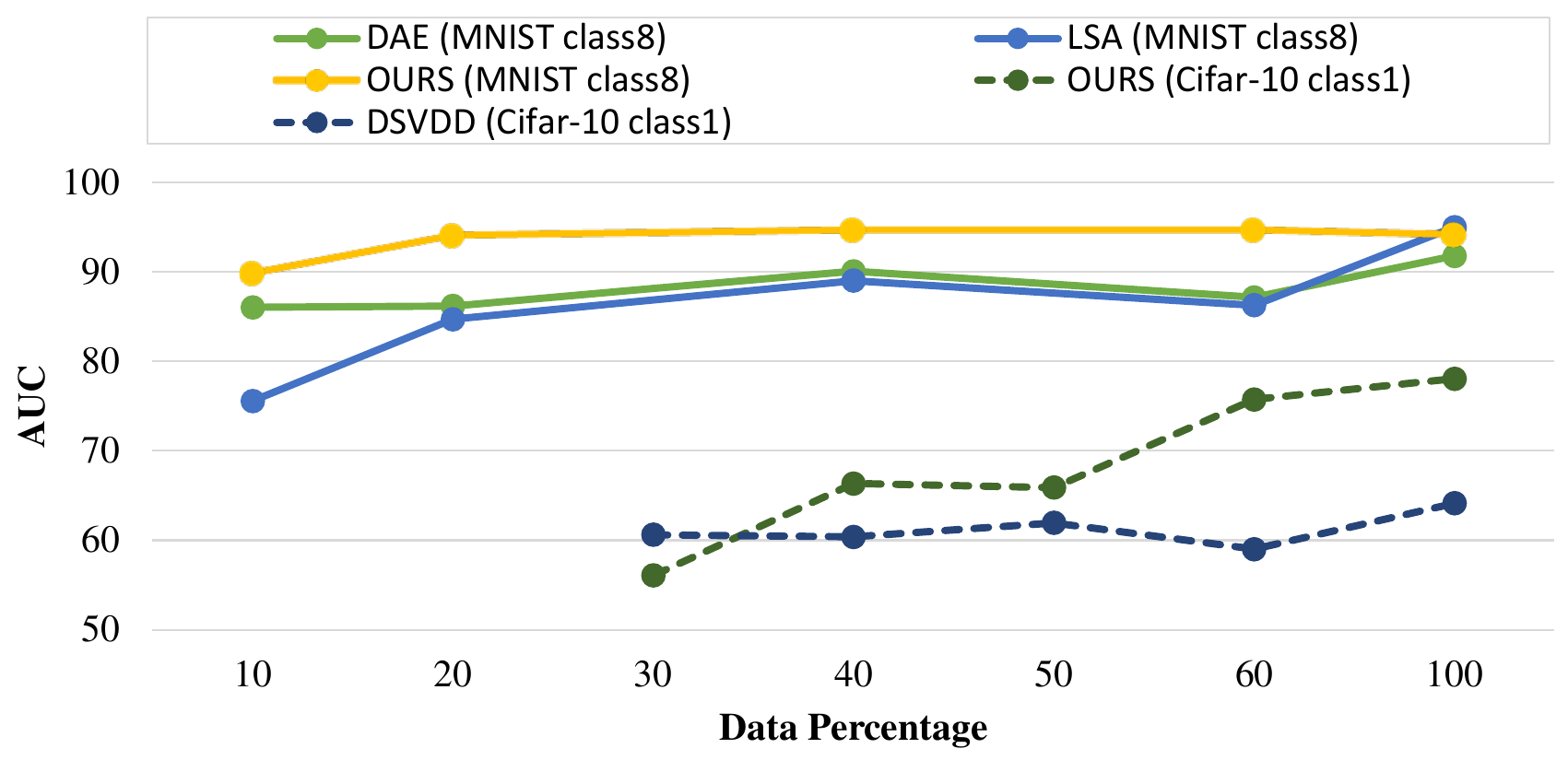}
      \caption{Puzzle-AE is significantly more data efficient than LSA \cite{abati2019latent} and DAE \cite{vincent2008extracting} on the class 8 of the MNIST \cite{lecun2010mnist} dataset. Puzzle-AE also performs better than DSVDD \cite{ruff2018deep} in the class car of the CIFAR-10 \cite{cifar} dataset.}
      \label{data_efficiency_plot}
\end{figure}

\textbf{Effect of training sample size on performance:} Data efficiency, as another important feature that is desired in real-world applications, is shown in Fig. \ref{data_efficiency_plot}. Traditional \ac{AE} based approaches usually need a  rich dataset to model every complexity in data and obtain good generalization. However, As Fig. \ref{data_efficiency_plot} illustrates, Puzzle-AE is significantly better than LSA \cite{abati2019latent} and DAE \cite{vincent2008extracting} in terms of data efficiency that is because of its different mean to model abstractions. It is also shown in table \ref{table:OURSvsLSA} that not only Puzzle-AE is significantly better than other SOTA AE-based approaches but also is better than DSVDD \cite{ruff2018deep} which is a one-class method.


\textbf{Robustness through attacked normal samples:} Robustness against attacked normal images was examined and the results are reported in Fig. \ref{plot:attack}. As it is shown, Puzzle-AE is significantly more robust against attacks to normal images with three different values of $\epsilon$ ($0.05, 0.1$ and $0.2$) in comparison with LSA \cite{abati2019latent}, ARAE \cite{salehi2020arae} and GT \cite{golan2018deep}. % and DAE\cite{vincent2008extracting}.
To explain different attacks applied to our method in attack1, we apply FGSM \cite{wong2020fast} attack on the normal class before permutation. However, in attack2, we apply FGSM \cite{wong2020fast} attack on the permuted image. The attacked image is brought back to the original form by using the corresponding inverse permutation. By averaging over all possible permutations, the attacked version of the normal class is obtained.  %GT\cite{golan2018deep}

\begin{figure*}[!ht]
\centering
     \includegraphics[width=\textwidth]{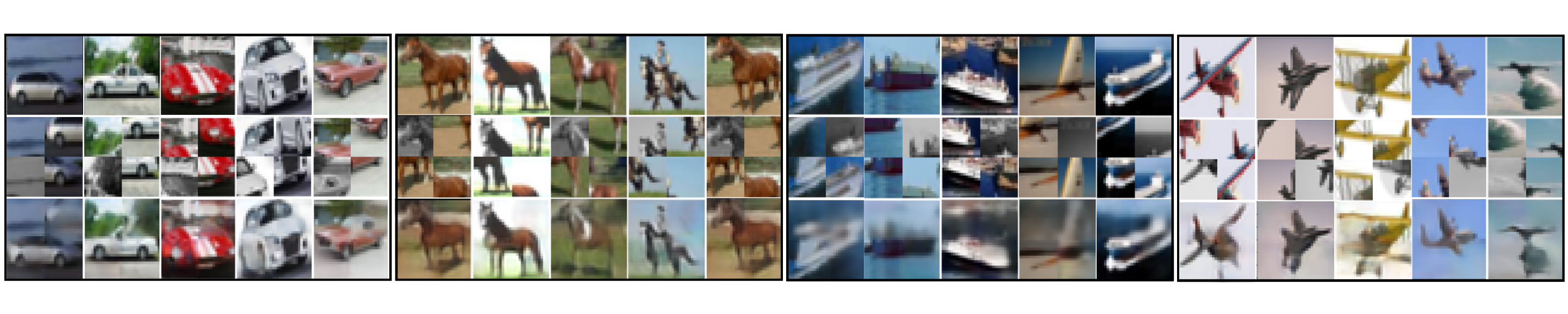}
      \caption{Visualization of the proposed method on CIFAR-10 \cite{cifar} dataset. First row is the original input, second row is the puzzled input mingled by colorization task and the third row is the unpuzzled output.}
      \label{Cifar_method}
\end{figure*}

\begin{figure}[!thb]
\centering
     \includegraphics[width=\linewidth]{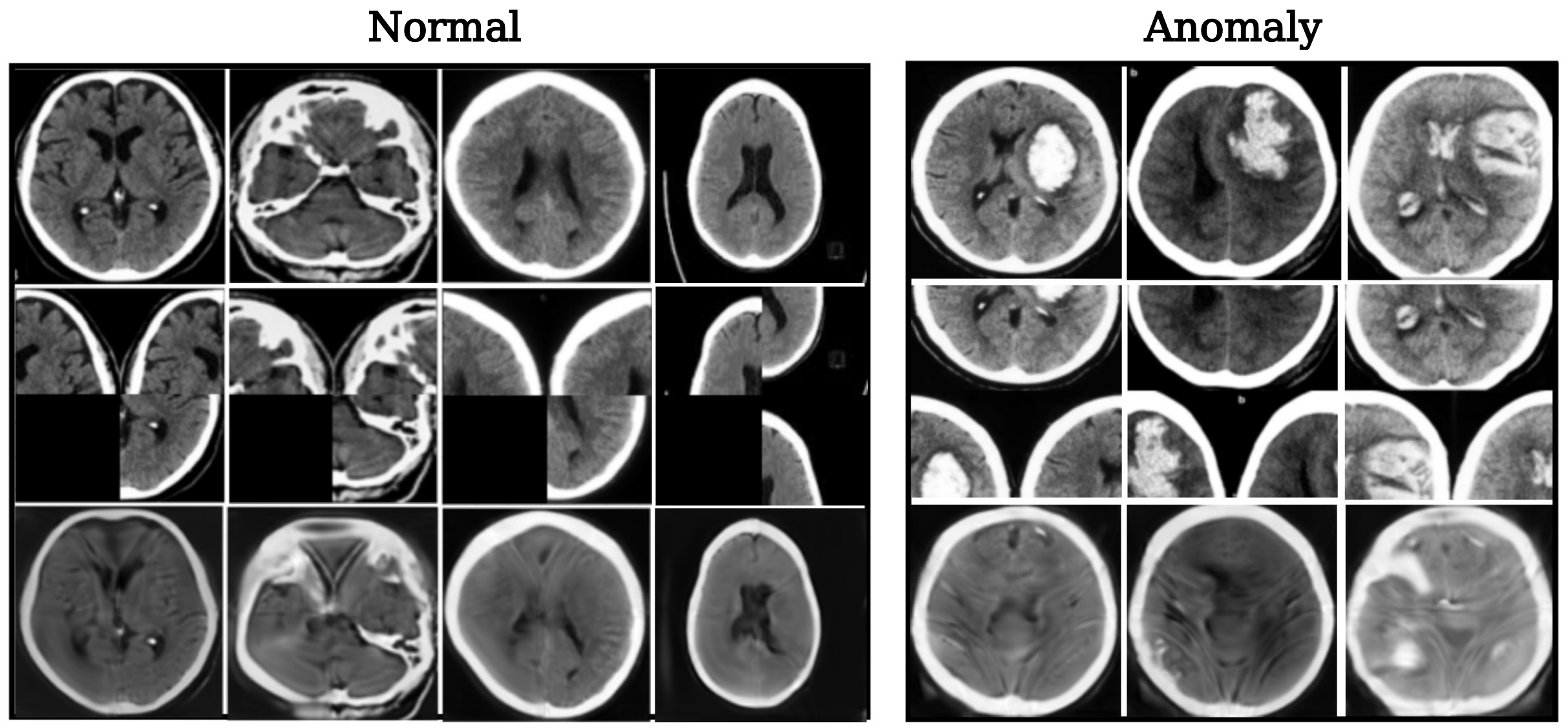}
      \caption{Visualization of the proposed method on Head CT dataset. First row is the original input, second row is the puzzled input mingled by the inpainting task, and the third row is the unpuzzled output solved by the model.}
      \label{medical_method}
\end{figure}

\textbf{Visualization of Puzzle-AE on Different Datasets:}
Sample images from different datasets are shown in Fig. \ref{Cifar_method}, \ref{Mnist_method}, and \ref{medical_method}. In each image, the first row shows the original input; the second row is the puzzled input mingled by one of the inpainting or colorization tasks. The final unpuzzled output of the model is shown in the third row.

% \vspace{2cm}
Fig. \ref{AnomalyReconstruction} shows outputs of the model trained on the class car of the CIFAR-10 \cite{cifar} dataset for some anomalous inputs.

% \textbf{Effect of each component of the proposed method on final performance:} Finally, we conduct experiments to evaluate the effect of every single module or algorithm that has been used in this framework on the CIFAR-10 \cite{cifar} dataset. The experiment and the results are provided in the appendix. As it can be seen, each of these parts has an important role and makes a significant contribution to gaining the best possible performance.
%  ----------------------Ablation----------------------
% GT fluctuation
\begin{table*}[ht]
\centering
\caption{Up - GT \cite{golan2018deep} AUC fluctuations during training procedure on head ct medical and the capsule class of  MVTecAD \cite{bergmann2019mvtec} dataset. Bottom - DSVDD \cite{ruff2018deep} AUC fluctuations during last 16 epochs of training procedure on the class car of CIFAR-10 \cite{cifar} dataset.}
\label{table:GT_epoch}
\resizebox{\textwidth}{!}{\begin{tabular}{ccccccccccccccccccccc}
\hline\noalign{\smallskip} 
Method & Dataset & & 0 & 1 & 2 & 3 & 4 & 5 & 6 & 7 & 8 & 9 & 10 & 11 & 12 & 13 & 14 & 15 & 16\\
\noalign{\smallskip}
\hline
\noalign{\smallskip}
\multirow{4}{*}{GT \cite{golan2018deep}}
& \multirow{2}{*}{Head CT} & AUC & 56.52 & 53.68 & 54.26 & 69.29 & 67.10 & 69.87 & 77.48 & 79.03 & \textbf{76.45} & 77.48 & 79.55 & 80.06 & \textbf{81.23} & 80.90 & 77.68 & 79.23 & \textbf{75.03}\\
% \noalign{\smallskip}
& & Train ACC & 14.59 & 34.45 & 43.41 & 55.84 & 71.17 & 90.08 & 98.88 & 99.92 & \textbf{99.98} & 100 & 100 & 100 & \textbf{100} & 100 & 99.93 & 99.60 & \textbf{98.14}\\
\noalign{\smallskip}
& \multirow{2}{*}{MVTec (Capsule)} & AUC & \textbf{45.95} & 67.41 & 71.20 & 74.11 & \textbf{77.46} & 60.75 & 64.10 & 73.00 & 73.00 & 62.82 & 62.50 & 66.29 & 69.49 & \textbf{60.63} & 66.85 & 73.35 & 68.93\\
% \noalign{\smallskip}
& & Train ACC & \textbf{98.89} & 99.50 & 100 & 100 & \textbf{100} & 99.99 & 100 & 100 & 100 & 100 & 100 & 99.55 & 100 & \textbf{98.49} & 100 & 100 & 100\\
\noalign{\smallskip}
\hline
\hline
\noalign{\smallskip}

\multirow{2}{*}{DSVDD \cite{ruff2018deep}}
& \multirow{2}{*}{Cifar-10 (Car)} & AUC & 56.21 & \textbf{51.83} & 53.74 & 54.50 & 57.57 & 59.86 & 60.23 & \textbf{60.60} & 59.81 & 56.37 & 57.72 & 55.82 & 58.40 & 54.04 & 54.97 & 55.25 & 56.22\\
% \noalign{\smallskip}
& & Train loss & 0.577 & \textbf{0.477} & 0.387 & 0.332 & 0.328 & 0.314 & 0.306 & \textbf{0.300} & 0.288 & 0.282 & 0.267 & 0.259 & 0.251 & 0.242 & 0.229 & 0.218 & 0.211\\
\noalign{\smallskip}
\hline
\noalign{\smallskip}

\end{tabular}}
\end{table*}

% \begin{figure}[!ht]
%     \centering
%     \parbox{\linewidth}{
%     \begin{subfigure}{\linewidth}
%             \includegraphics[width=\textwidth]{photos/plots/medical/head-ct-hemorrhage.pdf}
%     \end{subfigure}
    
%     \begin{subfigure}{\linewidth}
%             \includegraphics[width=\textwidth]{photos/plots/medical/brain_tumor_dataset.pdf}
           
%     \end{subfigure}\\
%     }
%     \caption{Min, max and avg \ac{AUC} Plots with respect to training epochs are shown for each medical dataset. As it is shown, training procedure continues till the convergence of all of the \ac{AUC} plots. }\label{medical_total}
% \end{figure}


\section{Ablation}
In this section, the effects of several important parameters are examined, as follows:
\begin{enumerate}[]
    \item \textbf{Effect of Each Component of the Puzzle-AE on Final Performance:} Table. \ref{table:min,max,avg} shows the effect of every single module or algorithm that has been used in this framework on the CIFAR-10 \cite{cifar} dataset. The results are shown for puzzle-solving AE (PAE), puzzle-solving and colorization (CPAE), and also CPAE combined with adversarial training (CPAE-G). The results show that each of these parts has an important role in gaining the best possible performance.
    % min, max, avg, all in one :))
\begin{table*}[ht]
\centering
\caption{Effect of each component or algorithm is illustrated separately. As it is shown, CPAE-G has the best results on sample dataset CIFAR-10 \cite{cifar}.}
\label{table:min,max,avg}
\resizebox{\textwidth}{!}{\begin{tabular}{ccccccccccccc}
\hline\noalign{\smallskip}
& & 0 & 1 & 2 & 3 & 4 & 5 & 6 & 7 & 8 & 9 & mean\\
\noalign{\smallskip}
\hline
\noalign{\smallskip}
\multirow{3}{*}{MIN} 
& puzzle AE (PAE) & 76.32 & 69.69 & 68.70 & 54.08 & 75.30 & 62.91 & 72.72 & 69.61 & 80.87 & 64.88 & 69.51 \\
& colorization + puzzle (CPAE) & 79.51 & 69.77 & 68.51 & 54.75 & 72.56 & 63.24 & 67.86 & 68.30 & 82.09 & 65.57 & 69.22 \\
& colorization + puzzle + GAN (CPAE-G) & 79.42 & 73.00 & 69.48 & 53.00 & 73.98 & 65.13 & 68.98 & 70.65 & 83.28 & 66.73 & 70.37 \\
\noalign{\smallskip}
\hline
\noalign{\smallskip}
\multirow{3}{*}{MAX} 
& puzzle AE (PAE) & 76.29 & 69.07 & 68.19 & 52.14 & 75.84 & 60.84 & 73.66 & 68.61 & 78.26 & 66.20 & 68.91 \\
& colorization + puzzle (CPAE) & 78.87 & 76.56 & 67.97 & 54.33 & 74.69 & 62.32 & 75.72 & 71.59 & 81.06 & 70.92 & 71.40 \\
& colorization + puzzle + GAN (CPAE-G) & 77.21 & 77.31 & 69.3 & 54.10 & 74.76 & 64.10 & 76.02 & 70.42 & 81.04 & 66.91 & 71.12 \\
\noalign{\smallskip}
\hline
\noalign{\smallskip}
\multirow{3}{*}{AVG} 
& puzzle AE (PAE) & 76.59 & 68.84 & 68.54 & 53.00 & 76.00 & 61.8 & 73.32 & 68.87 & 79.54 & 66.12 & 69.26 \\
& colorization + puzzle (CPAE) & 79.72 & 75.86 & 68.56 & 54.74 & 74.87 & 64.21 & 74.02 & 72.97 & 82.64 & 70.62 & 71.82 \\
& colorization + puzzle + GAN (CPAE-G) & 78.93 & 78.05 & 69.95 & 54.88 & 75.46 & 66.04 & 74.76 & 73.30 & 83.34 & 69.96 & \textbf{72.47}  \\
\noalign{\smallskip}
\hline
\noalign{\smallskip}

\end{tabular}}
\end{table*}
    
    \item \textbf{Effect of Puzzle Solving Compare to Rotation as the SSL Task:} As the Table. \ref{table:Other} reports, the performance of the rotation task is 1\% lesser than our method on the average of 15 classes. That is because of the rotation invariant aspect existing in most of the texture classes. As mentioned earlier, the puzzle-solving task is less ambiguous for different datasets than rotation prediction, which means we have lesser presumptions on training datasets for various problems.

    \item \textbf{Effect of converting pictures to gray-scale on performance:} It was observed that extracted normal features of Puzzle-AE are less sensitive to color for some specific classes of the CIFAR-10 \cite{cifar} dataset. For example, no significant performance drop was observed on the class car of CIFAR-10 \cite{cifar} when converting the whole dataset to gray-scale. Fig. \ref{grayscale} shows the ability of the proposed model in solving unseen gray-scale puzzled inputs. 
    
    % \item \textbf{Effect of changing activation functions on performance:}
    % In contrast to some of the \ac{SOTA} frameworks, it is observed that there is no significant performance drop when changing activation functions of our base model (standard U-Net\cite{li2018h}) to elu\cite{clevert2015fast} in Puzzle-AE. 
    
    \item \textbf{Effect of PGD \cite{madry2017towards},  FGSM \cite{wong2020fast}: } It was observed that PGD \cite{madry2017towards} and FGSM \cite{wong2020fast} have almost similar effects on the performance. They provide about 1\% improvement on datasets that have obvious shortcuts, such as MNIST. However, their role would be more significant in the robustification of the framework. 
    
    % \item \textbf{Effect of training sample size on performance: }It was observed that by dividing number of training samples by 12 in MNIST \cite{lecun2010mnist} dataset, the performance drop is about 3.1 percent. Table. \ref{table:OURSvsLSA} shows the results of the Puzzle-\ac{AE}, LSA \cite{abati2019latent} and DSVDD \cite{ruff2018deep} on MNIST \cite{lecun2010mnist} dataset with 1/12 of original training samples. 
    
\end{enumerate}

% 500data ours vs lsa
\begin{table*}[!ht]
\centering
\caption{
Puzzle-AE has ~6\% and ~3\% better performances when trained on the MNIST \cite{lecun2010mnist} with 1/12 samples with respect to LSA \cite{abati2019latent} and DSVDD \cite{ruff2018deep}.}
\label{table:OURSvsLSA}
\resizebox{\textwidth}{!}{\begin{tabular}{cccccccccccc}
\hline\noalign{\smallskip} 
Method & 0 & 1 & 2 & 3 & 4 & 5 & 6 & 7 & 8 & 9 & mean \\
\noalign{\smallskip}
\hline
\noalign{\smallskip}
LSA\textsuperscript{*} \cite{abati2019latent} & ${95.93\pm0.087}$ & ${99.82\pm0.005}$ & ${80.95\pm0.119}$ & ${83.95\pm0.090}$ & ${87.59\pm0.113}$ & ${85.81\pm0.072}$ & ${92.61\pm0.059}$ & ${93.31\pm0.074}$ & ${76.56\pm0.328}$ & ${92.40\pm0.075}$ & 88.89 \\
\noalign{\smallskip}
\hline
\noalign{\smallskip}
DSVDD\textsuperscript{*} \cite{ruff2018deep} & ${96.10\pm0.057}$ & ${99.17\pm0.006}$ & ${88.62\pm0.146}$ & ${86.60\pm0.252}$ & ${95.09\pm0.024}$ & ${84.93\pm0.091}$ & ${96.40\pm0.058}$ & ${94.68\pm0.036}$ & ${89.76\pm0.121}$ & ${94.72\pm0.044}$ & 92.61 \\
\noalign{\smallskip}
\hline
\noalign{\smallskip}
% max from ganomaly_gt_mnist_fmnist
OURS & ${99.54\pm0.044}$ & ${99.73\pm0.034}$ & ${90.41\pm0.593}$ & ${89.59\pm1.056}$ & ${95.71\pm0.306}$ & ${96.79\pm0.504}$ & ${97.23\pm0.310}$ & ${96.98\pm0.189}$ & ${89.81\pm0.676}$ & ${93.17\pm0.592}$ & \textbf{94.90}\\
\noalign{\smallskip}
\hline
% \noalign{\smallskip}

\end{tabular}}
\end{table*}

% FPR mnist table (ours vs lsa)
\begin{table*}[!t]
\centering
\caption{
Puzzle-AE has significantly better \ac{TPR}
at \ac{FPR} equal to 99.5\% and 99.0\% than LSA \cite{abati2019latent} on the MNIST \cite{lecun2010mnist} dataset for 10 class average.}
\label{table:FPR_MNIST}
\resizebox{0.98\textwidth}{!}
{\begin{tabular}{ccccccccccccc}
\hline\noalign{\smallskip} 
TPR & Method & 0 & 1 & 2 & 3 & 4 & 5 & 6 & 7 & 8 & 9 & Mean\\
\noalign{\smallskip}
\hline
\noalign{\smallskip}
\multirow{2}{*}{99\%}& LSA\textsuperscript{*} \cite{abati2019latent} & 0.0765 & 0.0088 & 0.7238 & 0.4119 & 0.6365 & 0.3274 & 0.1190 & 0.4591 & 0.5975 & 0.2428 & 0.3603\\
% from max
& OURS & 0.0480 & 0.0053 & 0.3295 & 0.2079 & 0.3157 & 0.1939 & 0.1086 & 0.2451 & 0.7156 & 0.1724 & \textbf{0.2342}\\
\noalign{\smallskip}
\hline
\noalign{\smallskip}
\multirow{2}{*}{99.5\%}& LSA\textsuperscript{*} \cite{abati2019latent} & 0.0980 & 0.0123 & 0.8178 & 0.4871 & 0.7210 & 0.3924 & 0.1587 & 0.5447 & 0.7146 & 0.3142 & 0.4261\\
% from max
& OURS & 0.0694 & 0.0088 & 0.4079 & 0.2545 & 0.4287 & 0.3262 & 0.1441 & 0.2977 & 0.7895 & 0.2061 & \textbf{0.2932}\\
\noalign{\smallskip}
\hline
% \noalign{\smallskip}

\end{tabular}}
\end{table*}

%  ----------------------Conclusion----------------
\section{Conclusion and Future Work}
In this study, we have tackled two significant problems of \ac{AE}s. Using a U-Net that solves puzzled inputs, the quality of reconstructed normal test time inputs is increased, and the inability to reconstruct anomalous samples is kept. The experimental results show significant improvements in stability, robustness, data efficiency, generality, and \ac{FPR} at high \ac{TPR}s  on a wide gamut of datasets. For the future works, some solutions for the several deficiencies of our method, such as solving some anomalous puzzled inputs, as is shown in the Fig. \ref{AnomalyReconstruction} will be investigated. Moreover, a quantitative criterion for selecting min, max, or the average of the anomaly score could improve the results for future works.

\ifCLASSOPTIONcompsoc
  % The Computer Society usually uses the plural form
  \section*{Acknowledgments}
\else
  % regular IEEE prefers the singular form
  \section*{Acknowledgment}
\fi

The authors would like to thank Soroosh Baselizadeh and Amirreza Shaeiri for their insightful comments and reviews.
\\

% Can use something like this to put references on a page
% by themselves when using endfloat and the captionsoff option.
\ifCLASSOPTIONcaptionsoff
  \newpage
\fi

% trigger a \newpage just before the given reference
% number - used to balance the columns on the last page
% adjust value as needed - may need to be readjusted if
% the document is modified later
%\IEEEtriggeratref{8}
% The "triggered" command can be changed if desired:
%\IEEEtriggercmd{\enlargethispage{-5in}}

% references section

% can use a bibliography generated by BibTeX as a .bbl file
% BibTeX documentation can be easily obtained at:
% http://mirror.ctan.org/biblio/bibtex/contrib/doc/
% The IEEEtran BibTeX style support page is at:
% http://www.michaelshell.org/tex/ieeetran/bibtex/

% \newpage
\clearpage
\bibliographystyle{IEEEtran}
% argument is your BibTeX string definitions and bibliography database(s)
\bibliography{references}

% Generated by IEEEtran.bst, version: 1.14 (2015/08/26)
\begin{thebibliography}{10}
\providecommand{\url}[1]{#1}
\csname url@samestyle\endcsname
\providecommand{\newblock}{\relax}
\providecommand{\bibinfo}[2]{#2}
\providecommand{\BIBentrySTDinterwordspacing}{\spaceskip=0pt\relax}
\providecommand{\BIBentryALTinterwordstretchfactor}{4}
\providecommand{\BIBentryALTinterwordspacing}{\spaceskip=\fontdimen2\font plus
\BIBentryALTinterwordstretchfactor\fontdimen3\font minus
  \fontdimen4\font\relax}
\providecommand{\BIBforeignlanguage}[2]{{%
\expandafter\ifx\csname l@#1\endcsname\relax
\typeout{** WARNING: IEEEtran.bst: No hyphenation pattern has been}%
\typeout{** loaded for the language `#1'. Using the pattern for}%
\typeout{** the default language instead.}%
\else
\language=\csname l@#1\endcsname
\fi
#2}}
\providecommand{\BIBdecl}{\relax}
\BIBdecl

\bibitem{chalapathy2019deep}
R.~Chalapathy and S.~Chawla, ``Deep learning for anomaly detection: A survey,''
  \emph{arXiv preprint arXiv:1901.03407}, 2019.

\bibitem{chen2018unsupervised}
X.~Chen and E.~Konukoglu, ``Unsupervised detection of lesions in brain mri
  using constrained adversarial auto-encoders,'' \emph{arXiv preprint
  arXiv:1806.04972}, 2018.

\bibitem{baur2018deep}
C.~Baur, B.~Wiestler, S.~Albarqouni, and N.~Navab, ``Deep autoencoding models
  for unsupervised anomaly segmentation in brain mr images,'' in
  \emph{International MICCAI Brainlesion Workshop}.\hskip 1em plus 0.5em minus
  0.4em\relax Springer, 2018, pp. 161--169.

\bibitem{gong2019memorizing}
D.~Gong, L.~Liu, V.~Le, B.~Saha, M.~R. Mansour, S.~Venkatesh, and A.~v.~d.
  Hengel, ``Memorizing normality to detect anomaly: Memory-augmented deep
  autoencoder for unsupervised anomaly detection,'' in \emph{Proceedings of the
  IEEE International Conference on Computer Vision}, 2019, pp. 1705--1714.

\bibitem{goodfellow2016nips}
I.~Goodfellow, ``Nips 2016 tutorial: Generative adversarial networks,''
  \emph{arXiv preprint arXiv:1701.00160}, 2016.

\bibitem{kodali2017convergence}
N.~Kodali, J.~Abernethy, J.~Hays, and Z.~Kira, ``On convergence and stability
  of gans,'' \emph{arXiv preprint arXiv:1705.07215}, 2017.

\bibitem{salimans2016improved}
T.~Salimans, I.~Goodfellow, W.~Zaremba, V.~Cheung, A.~Radford, and X.~Chen,
  ``Improved techniques for training gans,'' in \emph{Advances in neural
  information processing systems}, 2016, pp. 2234--2242.

\bibitem{martin2017towards}
A.~Martin and B.~Lon, ``Towards principled methods for training generative
  adversarial networks,'' in \emph{NIPS 2016 Workshop on Adversarial Training.
  In review for ICLR}, vol. 2016, 2017.

\bibitem{nalisnick2018deep}
E.~Nalisnick, A.~Matsukawa, Y.~W. Teh, D.~Gorur, and B.~Lakshminarayanan, ``Do
  deep generative models know what they don't know?'' \emph{arXiv preprint
  arXiv:1810.09136}, 2018.

\bibitem{ruff2018deep}
L.~Ruff, R.~Vandermeulen, N.~Goernitz, L.~Deecke, S.~A. Siddiqui, A.~Binder,
  E.~M{\"u}ller, and M.~Kloft, ``Deep one-class classification,'' in
  \emph{International conference on machine learning}, 2018, pp. 4393--4402.

\bibitem{sakurada2014anomaly}
M.~Sakurada and T.~Yairi, ``Anomaly detection using autoencoders with nonlinear
  dimensionality reduction,'' in \emph{Proceedings of the MLSDA 2014 2nd
  Workshop on Machine Learning for Sensory Data Analysis}, 2014, pp. 4--11.

\bibitem{cifar}
A.~Krizhevsky, V.~Nair, and G.~Hinton, ``Cifar-10 (canadian institute for
  advanced research),'' 2009.

\bibitem{deng2009imagenet}
J.~Deng, W.~Dong, R.~Socher, L.-J. Li, K.~Li, and L.~Fei-Fei, ``Imagenet: A
  large-scale hierarchical image database,'' in \emph{2009 IEEE conference on
  computer vision and pattern recognition}.\hskip 1em plus 0.5em minus
  0.4em\relax Ieee, 2009, pp. 248--255.

\bibitem{wu2018unsupervised}
Z.~Wu, Y.~Xiong, S.~X. Yu, and D.~Lin, ``Unsupervised feature learning via
  non-parametric instance discrimination,'' in \emph{Proceedings of the IEEE
  Conference on Computer Vision and Pattern Recognition}, 2018, pp. 3733--3742.

\bibitem{oord2018representation}
A.~v.~d. Oord, Y.~Li, and O.~Vinyals, ``Representation learning with
  contrastive predictive coding,'' \emph{arXiv preprint arXiv:1807.03748},
  2018.

\bibitem{zhang2017split}
R.~Zhang, P.~Isola, and A.~A. Efros, ``Split-brain autoencoders: Unsupervised
  learning by cross-channel prediction,'' in \emph{Proceedings of the IEEE
  Conference on Computer Vision and Pattern Recognition}, 2017, pp. 1058--1067.

\bibitem{golan2018deep}
I.~Golan and R.~El-Yaniv, ``Deep anomaly detection using geometric
  transformations,'' in \emph{Advances in Neural Information Processing
  Systems}, 2018, pp. 9758--9769.

\bibitem{bergmann2019mvtec}
P.~Bergmann, M.~Fauser, D.~Sattlegger, and C.~Steger, ``Mvtec ad--a
  comprehensive real-world dataset for unsupervised anomaly detection,'' in
  \emph{Proceedings of the IEEE Conference on Computer Vision and Pattern
  Recognition}, 2019, pp. 9592--9600.

\bibitem{kitamura2018hemorrhage}
F.~Kitamura, ``Head ct - hemorrhage,''
  \url{https://www.kaggle.com/felipekitamura/head-ct-hemorrhage}, 2018.

\bibitem{chakrabarty2019tumor}
N.~Chakrabarty, ``Brain mri images for brain tumor detection,''
  \url{https://www.kaggle.com/navoneel/brain-mri-images-for-brain-tumor-detection},
  2019.

\bibitem{ronneberger2015u}
O.~Ronneberger, P.~Fischer, and T.~Brox, ``U-net: Convolutional networks for
  biomedical image segmentation,'' in \emph{International Conference on Medical
  image computing and computer-assisted intervention}.\hskip 1em plus 0.5em
  minus 0.4em\relax Springer, 2015, pp. 234--241.

\bibitem{badrinarayanan2017segnet}
V.~Badrinarayanan, A.~Kendall, and R.~Cipolla, ``Segnet: A deep convolutional
  encoder-decoder architecture for image segmentation,'' \emph{IEEE
  transactions on pattern analysis and machine intelligence}, vol.~39, no.~12,
  pp. 2481--2495, 2017.

\bibitem{litjens2017survey}
G.~Litjens, T.~Kooi, B.~E. Bejnordi, A.~A.~A. Setio, F.~Ciompi, M.~Ghafoorian,
  J.~A. Van Der~Laak, B.~Van~Ginneken, and C.~I. S{\'a}nchez, ``A survey on
  deep learning in medical image analysis,'' \emph{Medical image analysis},
  vol.~42, pp. 60--88, 2017.

\bibitem{noroozi2016unsupervised}
M.~Noroozi and P.~Favaro, ``Unsupervised learning of visual representations by
  solving jigsaw puzzles,'' in \emph{European Conference on Computer
  Vision}.\hskip 1em plus 0.5em minus 0.4em\relax Springer, 2016, pp. 69--84.

\bibitem{hendrycks2019using}
D.~Hendrycks, M.~Mazeika, S.~Kadavath, and D.~Song, ``Using self-supervised
  learning can improve model robustness and uncertainty,'' in \emph{Advances in
  Neural Information Processing Systems}, 2019, pp. 15\,663--15\,674.

\bibitem{madry2017towards}
A.~Madry, A.~Makelov, L.~Schmidt, D.~Tsipras, and A.~Vladu, ``Towards deep
  learning models resistant to adversarial attacks,'' \emph{arXiv preprint
  arXiv:1706.06083}, 2017.

\bibitem{ilyas2019adversarial}
A.~Ilyas, S.~Santurkar, D.~Tsipras, L.~Engstrom, B.~Tran, and A.~Madry,
  ``Adversarial examples are not bugs, they are features,'' in \emph{Advances
  in Neural Information Processing Systems}, 2019, pp. 125--136.

\bibitem{sabokrou2018adversarially}
M.~Sabokrou, M.~Khalooei, M.~Fathy, and E.~Adeli, ``Adversarially learned
  one-class classifier for novelty detection,'' in \emph{Proceedings of the
  IEEE Conference on Computer Vision and Pattern Recognition}, 2018, pp.
  3379--3388.

\bibitem{abati2019latent}
D.~Abati, A.~Porrello, S.~Calderara, and R.~Cucchiara, ``Latent space
  autoregression for novelty detection,'' in \emph{Proceedings of the IEEE
  Conference on Computer Vision and Pattern Recognition}, 2019, pp. 481--490.

\bibitem{perera2019ocgan}
P.~Perera, R.~Nallapati, and B.~Xiang, ``Ocgan: One-class novelty detection
  using gans with constrained latent representations,'' in \emph{Proceedings of
  the IEEE Conference on Computer Vision and Pattern Recognition}, 2019, pp.
  2898--2906.

\bibitem{schlegl2017unsupervised}
T.~Schlegl, P.~Seeb{\"o}ck, S.~M. Waldstein, U.~Schmidt-Erfurth, and G.~Langs,
  ``Unsupervised anomaly detection with generative adversarial networks to
  guide marker discovery,'' in \emph{International conference on information
  processing in medical imaging}.\hskip 1em plus 0.5em minus 0.4em\relax
  Springer, 2017, pp. 146--157.

\bibitem{akcay2018ganomaly}
S.~Akcay, A.~Atapour-Abarghouei, and T.~P. Breckon, ``Ganomaly: Semi-supervised
  anomaly detection via adversarial training,'' in \emph{Asian Conference on
  Computer Vision}.\hskip 1em plus 0.5em minus 0.4em\relax Springer, 2018, pp.
  622--637.

\bibitem{kingma2013auto}
D.~P. Kingma and M.~Welling, ``Auto-encoding variational bayes,'' \emph{arXiv
  preprint arXiv:1312.6114}, 2013.

\bibitem{goodfellow2014explaining}
I.~J. Goodfellow, J.~Shlens, and C.~Szegedy, ``Explaining and harnessing
  adversarial examples,'' \emph{arXiv preprint arXiv:1412.6572}, 2014.

\bibitem{li2018h}
X.~Li, H.~Chen, X.~Qi, Q.~Dou, C.-W. Fu, and P.-A. Heng, ``H-denseunet: hybrid
  densely connected unet for liver and tumor segmentation from ct volumes,''
  \emph{IEEE transactions on medical imaging}, vol.~37, no.~12, pp. 2663--2674,
  2018.

\bibitem{wang2017chestx}
X.~Wang, Y.~Peng, L.~Lu, Z.~Lu, M.~Bagheri, and R.~M. Summers, ``Chestx-ray8:
  Hospital-scale chest x-ray database and benchmarks on weakly-supervised
  classification and localization of common thorax diseases,'' in
  \emph{Proceedings of the IEEE conference on computer vision and pattern
  recognition}, 2017, pp. 2097--2106.

\bibitem{salehi2020arae}
M.~Salehi, A.~Arya, B.~Pajoum, M.~Otoofi, A.~Shaeiri, M.~H. Rohban, and H.~R.
  Rabiee, ``Arae: Adversarially robust training of autoencoders improves
  novelty detection,'' \emph{arXiv preprint arXiv:2003.05669}, 2020.

\bibitem{wong2020fast}
E.~Wong, L.~Rice, and J.~Z. Kolter, ``Fast is better than free: Revisiting
  adversarial training,'' \emph{arXiv preprint arXiv:2001.03994}, 2020.

\bibitem{sabokrou2018avid}
M.~Sabokrou, M.~Pourreza, M.~Fayyaz, R.~Entezari, M.~Fathy, J.~Gall, and
  E.~Adeli, ``Avid: Adversarial visual irregularity detection,'' in \emph{Asian
  Conference on Computer Vision}.\hskip 1em plus 0.5em minus 0.4em\relax
  Springer, 2018, pp. 488--505.

\bibitem{tolstikhin2017wasserstein}
I.~Tolstikhin, O.~Bousquet, S.~Gelly, and B.~Schoelkopf, ``Wasserstein
  auto-encoders,'' \emph{arXiv preprint arXiv:1711.01558}, 2017.

\bibitem{lecun2010mnist}
Y.~LeCun, C.~Cortes, and C.~Burges, ``Mnist handwritten digit database,'' 2010.

\bibitem{chen2001one}
Y.~Chen, X.~S. Zhou, and T.~S. Huang, ``One-class svm for learning in image
  retrieval,'' in \emph{Proceedings 2001 International Conference on Image
  Processing (Cat. No. 01CH37205)}, vol.~1.\hskip 1em plus 0.5em minus
  0.4em\relax IEEE, 2001, pp. 34--37.

\bibitem{li2019icml}
X.~Li, I.~Kiringa, T.~Yeap, X.~Zhu, and Y.~Li, ``Exploring deep anomaly
  detection methods based on capsule net,'' in \emph{ICML 2019 Workshop on
  Uncertainty and Robustness in Deep Learning, At Long Beach}, 2019.

\bibitem{xiao2017fashion}
H.~Xiao, K.~Rasul, and R.~Vollgraf, ``Fashion-mnist: a novel image dataset for
  benchmarking machine learning algorithms,'' \emph{arXiv preprint
  arXiv:1708.07747}, 2017.

\bibitem{zong2018deep}
B.~Zong, Q.~Song, M.~R. Min, W.~Cheng, C.~Lumezanu, D.~Cho, and H.~Chen, ``Deep
  autoencoding gaussian mixture model for unsupervised anomaly detection,''
  2018.

\bibitem{zhai2016deep}
S.~Zhai, Y.~Cheng, W.~Lu, and Z.~Zhang, ``Deep structured energy based models
  for anomaly detection,'' \emph{arXiv preprint arXiv:1605.07717}, 2016.

\bibitem{bergmann2019visigrapp}
P.~Bergmann, S.~L\"{o}we, M.~Fauser, D.~Sattlegger, and C.~Steger, ``Improving
  unsupervised defect segmentation by applying structural similarity to
  autoencoders,'' in \emph{International Joint Conference on Computer Vision,
  Imaging and Computer Graphics Theory and Applications (VISIGRAPP)}, 2019.

\bibitem{nene1996columbia}
S.~A. Nene, S.~K. Nayar, H.~Murase \emph{et~al.}, ``Columbia object image
  library (coil-100),'' 1996.

\bibitem{ioffe2015batch}
S.~Ioffe and C.~Szegedy, ``Batch normalization: Accelerating deep network
  training by reducing internal covariate shift,'' \emph{arXiv preprint
  arXiv:1502.03167}, 2015.

\bibitem{radford2015unsupervised}
A.~Radford, L.~Metz, and S.~Chintala, ``Unsupervised representation learning
  with deep convolutional generative adversarial networks,'' \emph{arXiv
  preprint arXiv:1511.06434}, 2015.

\bibitem{kingma2014adam}
D.~P. Kingma and J.~Ba, ``Adam: A method for stochastic optimization,''
  \emph{arXiv preprint arXiv:1412.6980}, 2014.

\bibitem{goyal2017accurate}
P.~Goyal, P.~Doll{\'a}r, R.~Girshick, P.~Noordhuis, L.~Wesolowski, A.~Kyrola,
  A.~Tulloch, Y.~Jia, and K.~He, ``Accurate, large minibatch sgd: Training
  imagenet in 1 hour,'' \emph{arXiv preprint arXiv:1706.02677}, 2017.

\bibitem{Nene96objectimage}
S.~A. Nene, S.~K. Nayar, and H.~Murase, ``object image library (coil-100,''
  Tech. Rep., 1996.

\bibitem{xu2010robust}
H.~Xu, C.~Caramanis, and S.~Sanghavi, ``Robust pca via outlier pursuit,'' in
  \emph{Advances in Neural Information Processing Systems}, 2010, pp.
  2496--2504.

\bibitem{tsakiris2018dual}
M.~C. Tsakiris and R.~Vidal, ``Dual principal component pursuit,'' \emph{The
  Journal of Machine Learning Research}, vol.~19, no.~1, pp. 684--732, 2018.

\bibitem{pidhorskyi2018generative}
S.~Pidhorskyi, R.~Almohsen, and G.~Doretto, ``Generative probabilistic novelty
  detection with adversarial autoencoders,'' in \emph{Advances in neural
  information processing systems}, 2018, pp. 6822--6833.

\bibitem{gidaris2018unsupervised}
S.~Gidaris, P.~Singh, and N.~Komodakis, ``Unsupervised representation learning
  by predicting image rotations,'' \emph{arXiv preprint arXiv:1803.07728},
  2018.

\bibitem{chen2018convergence}
X.~Chen, S.~Liu, R.~Sun, and M.~Hong, ``On the convergence of a class of
  adam-type algorithms for non-convex optimization,'' \emph{arXiv preprint
  arXiv:1808.02941}, 2018.

\bibitem{vincent2008extracting}
P.~Vincent, H.~Larochelle, Y.~Bengio, and P.-A. Manzagol, ``Extracting and
  composing robust features with denoising autoencoders,'' in \emph{Proceedings
  of the 25th international conference on Machine learning}, 2008, pp.
  1096--1103.

\end{thebibliography}

\begin{IEEEbiography}[{\includegraphics[width=1in,height=1.25in,clip,keepaspectratio]{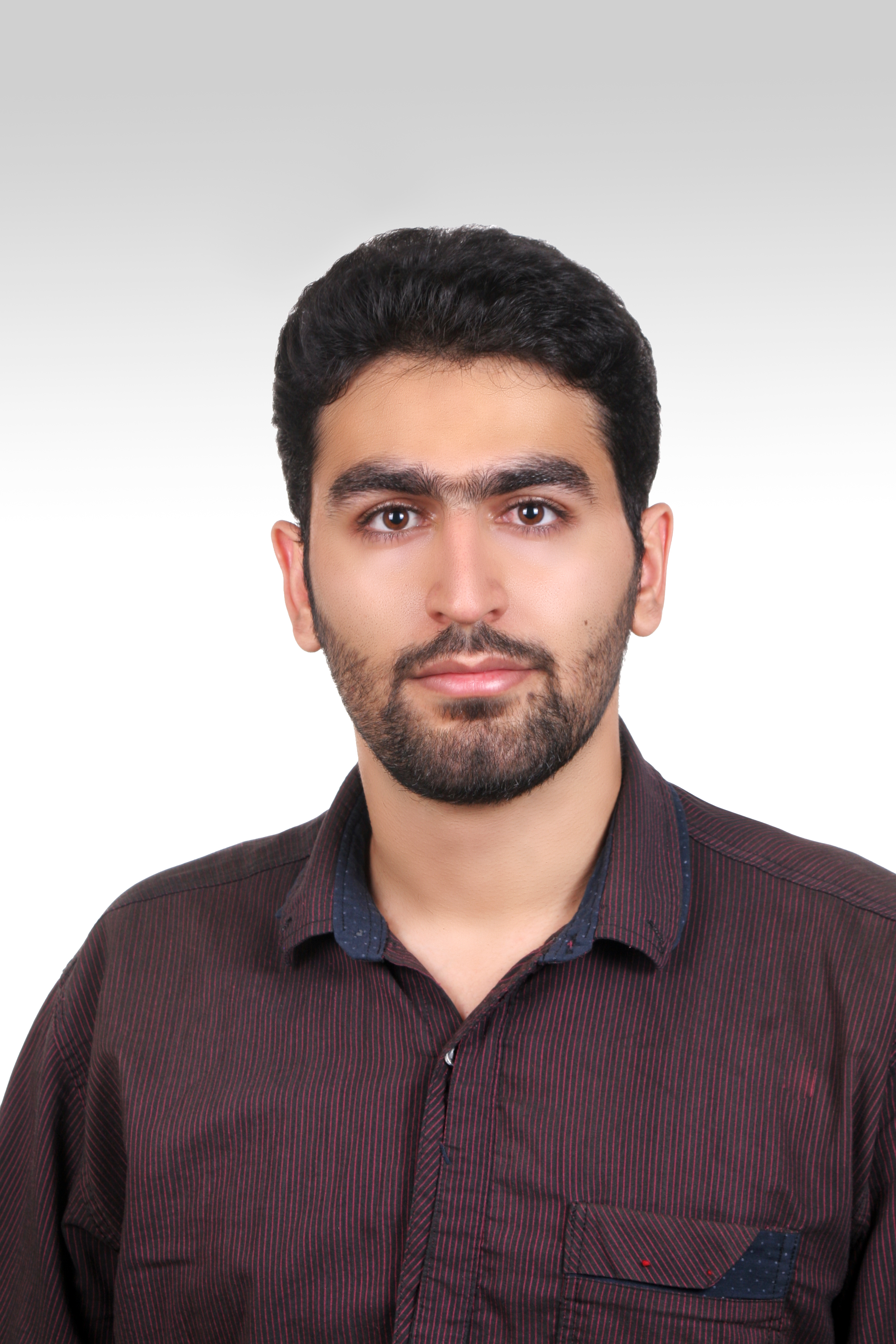}}]{Mohammadreza Salehi}
received his B.Sc. degree in computer engineering from the University of Tehran in 2018. He is currently an M.Sc. student in the Department of Computer Engineering, Sharif University of Technology. His research interests include computer vision and machine learning, focusing on anomaly detection in image and video.
\end{IEEEbiography}

\begin{IEEEbiography}[{\includegraphics[width=1in,height=1.25in,clip,keepaspectratio]{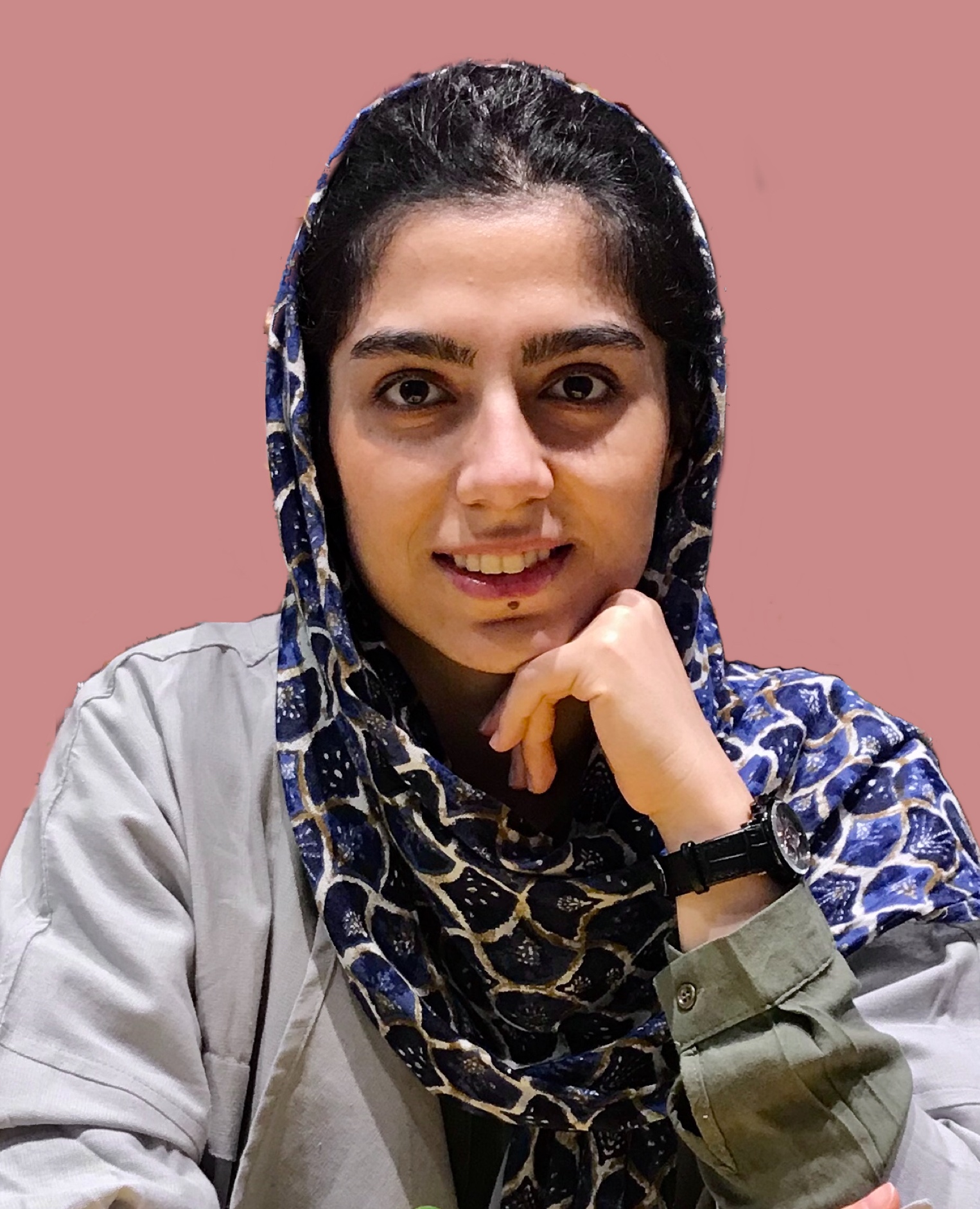}}]{Ainaz Eftekhar}
is currently pursuing her B.Sc. degree in the Department of Computer Engineering at the Sharif University of Technology. She is working as a researcher in the Robust and Interpretable ML Lab and as a remote research assistant in Visual Intelligence and Learning Laboratory (VILAB) at EPFL. Her research interests include computer vision, machine learning, and AI.
\end{IEEEbiography}

\begin{IEEEbiography}[{\includegraphics[width=1in,height=1.25in,clip,keepaspectratio]{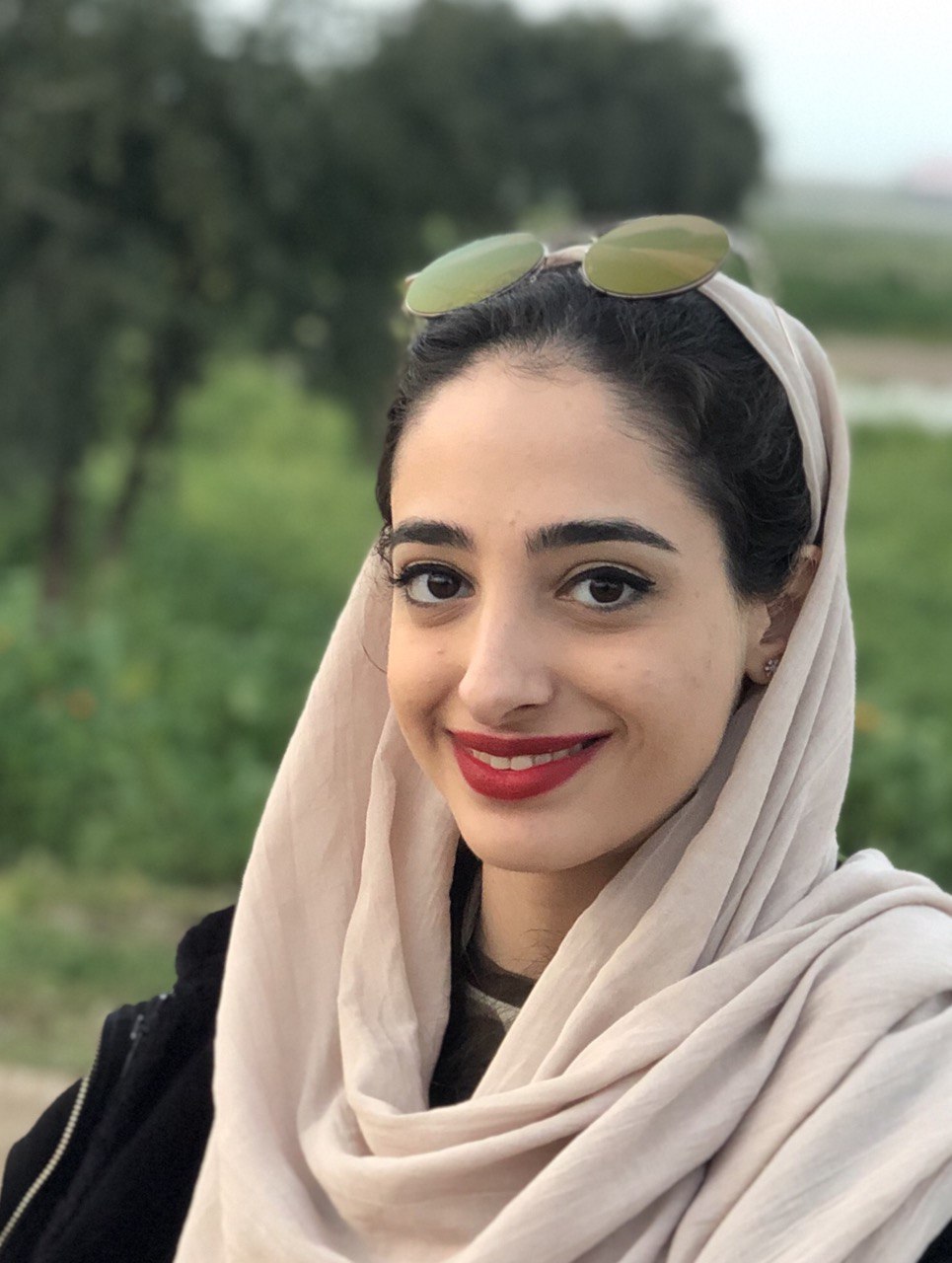}}]{Niousha Sadjadi}
is a senior undergraduate student in the Department of Computer Engineering at the Sharif University of Technology. Currently, she works as a researcher on different anomaly detection methods in the Robust and Interpretable ML lab. Her research interests include computer vision, unsupervised learning, neural computing, and anomaly detection on images.
\end{IEEEbiography}

\begin{IEEEbiography}[{\includegraphics[width=1in,height=1.25in,clip,keepaspectratio]{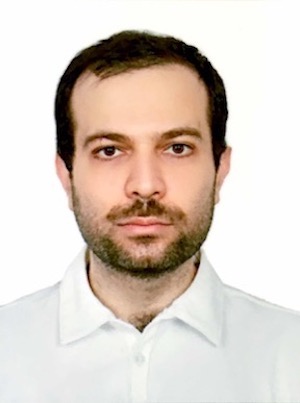}}]{Mohammad Hossein Rohban}
received his BS, MS ad Ph.D. degrees in Computer Engineering from the Sharif University of Technology. Currently, he is an assistant professor in the Department of Computer Engineering at the Sharif University of Technology. His current research interests include interpretable and robust machine learning, anomaly detection, and computational Biology. He previously spent three years as a postdoctoral associate at the Broad Institute of Harvard and MIT. He focused on various problems at the intersection of machine learning and image-based Computational Biology, where he published several prestigious papers on the mentioned subjects in the relevant journals.
\end{IEEEbiography}

\begin{IEEEbiography}[{\includegraphics[width=1in,height=1.25in,clip,keepaspectratio]{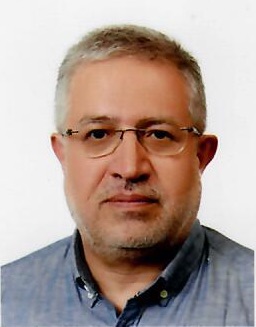}}]{Hamid R. Rabiee}
(SM ’07) received his BS and MS degrees (with Great Distinction) in Electrical Engineering from CSULB, Long Beach, CA (1987, 1989), his EEE degree in Electrical and Computer Engineering from USC, Los Angeles, CA (1993), and his Ph.D. in Electrical and Computer Engineering from Purdue University, West Lafayette, IN, in 1996. From 1993 to 1996, he was a Member of the Technical Staff at AT\&T Bell Laboratories. From 1996 to 1999, he worked as a Senior Software Engineer at Intel Corporation. He was also with PSU, OGI, and OSU universities as an adjunct professor of Electrical and Computer Engineering from 1996-2000. Since September 2000, he has joined the Sharif University of Technology, Tehran, Iran. He was also a visiting professor at the Imperial College of London for the 2017-2018 academic year. He is the founder of Sharif University Advanced Information and Communication Technology Research Institute (AICT), ICT Innovation Center, Advanced Technologies Incubator (SATI), Digital Media Laboratory (DML), Mobile Value Added Services Laboratory (VASL), Bioinformatics and Computational Biology Laboratory (BCB) and Cognitive Neuroengineering Research Center. He is also a consultant and member of AI in Health Expert Group at WHO. He has been the founder of many successful High-Tech start-up companies in ICT as an entrepreneur. He is currently a Professor of Computer Engineering at Sharif University of Technology, and Director of AICT, DML, and VASL. He has received numerous awards and
honors for his Industrial, scientific, and academic contributions and holds three patents. His research interests include statistical machine learning, Bayesian statistics, data analytics, and complex networks with applications in social networks, multimedia systems, cloud and IoT privacy, bioinformatics, and brain networks.
\end{IEEEbiography}

% if you will not have a photo at all:
% \begin{IEEEbiographynophoto}{John Doe}
% Biography text here.
% \end{IEEEbiographynophoto}

% insert where needed to balance the two columns on the last page with
% biographies
%\newpage

% \begin{IEEEbiographynophoto}{Jane Doe}
% Biography text here.
% \end{IEEEbiographynophoto}

% You can push biographies down or up by placing
% a \vfill before or after them. The appropriate
% use of \vfill depends on what kind of text is
% on the last page and whether or not the columns
% are being equalized.

%\vfill

% Can be used to pull up biographies so that the bottom of the last one
% is flush with the other column.
%\enlargethispage{-5in}

% that's all folks
\end{document}